%% file: nips_2016.tex
\documentclass{article}
\pdfoutput=1
\PassOptionsToPackage{numbers, compress}{natbib}
\usepackage[final]{nips_2016}

\usepackage[utf8]{inputenc} % allow utf-8 input
\usepackage[T1]{fontenc}    % use 8-bit T1 fonts
\usepackage{url}            % simple URL typesetting
\usepackage{booktabs}       % professional-quality tables
\usepackage{amsfonts}       % blackboard math symbols
\usepackage{nicefrac}       % compact symbols for 1/2, etc.
\usepackage{microtype}      % microtypography
\usepackage{graphicx}
\usepackage{amsmath, amssymb, amsthm}
\usepackage{multirow}
\usepackage{wrapfig}
\usepackage{sidecap}
\usepackage{authblk}

\title{Value Iteration Networks}

\author{Aviv Tamar}
\author{Yi Wu}
\author{Garrett Thomas}
\author{Sergey Levine}
\author{Pieter Abbeel}
  
\affil{Dept. of Electrical Engineering and Computer Sciences, UC Berkeley}

\input{defs}

\begin{document}

\maketitle

\begin{abstract}
We introduce the \emph{value iteration network} (VIN): a fully differentiable neural network with a `planning module' embedded within. VINs can \emph{learn to plan}, and are suitable for predicting outcomes that involve planning-based reasoning, such as policies for reinforcement learning. Key to our approach is a novel \emph{differentiable} approximation of the value-iteration algorithm, which can be represented as a convolutional neural network, and trained end-to-end using standard backpropagation.
We evaluate VIN based policies on discrete and continuous path-planning domains, and on a natural-language based search task. We show that by learning an explicit planning computation, VIN policies generalize better to new, unseen domains.
\end{abstract}

\section{Introduction}\label{s:intro}
\input{intro}
\vspace{-1em}
\section{Background}\label{s:background}
\vspace{-0.5em}
\input{background}

\vspace{-1em}
\section{The Value Iteration Network Model}\label{s:VIN}
\vspace{-10pt}
\input{model}

\vspace{-10pt}
\section{Experiments}\label{s:experiments}
\vspace{-10pt}
In this section we evaluate VINs as policy representations on various domains. Additional experiments investigating RL and hierarchical VINs, as well as technical implementation details are discussed in the supplementary material. Source code is available at \url{https://github.com/avivt/VIN}.
\\
Our goal in these experiments is to investigate the following questions:
\vspace{-5pt}
\begin{enumerate}
\item Can VINs effectively learn a planning computation using standard RL and IL algorithms? 
%\item Can end-to-end training effectively produce non-trivial VIN policies?
\item Does the planning computation learned by VINs make them better than reactive policies at generalizing to new domains?
\end{enumerate}
\vspace{-5pt}
An additional goal is to point out several ideas for designing VINs for various tasks. While this is not an exhaustive list that fits all domains, we hope that it will motivate creative designs in future work.
% Due to lack of space, we report some of the technical implementation details in the supplementary material. 

\vspace{-10pt}
\subsection{Grid-World Domain}\label{ss:grid-world}
\vspace{-5pt}
\input{gridworld_experiments}

%We evaluated the VIN model in two path-planning domains. The first domain is a synthetic grid-world with obstacles, in which the observation is a map of the obstacles and the goal position. The second domain models a navigation task for a Mars rover. In this domain, the observation is an overhead image of the terrain, along with a map of the goal position. 
%Our goal in these experiments is two-fold. First, we wish to show that VIN models are vastly superior to standard feed-forward networks in prediction tasks that involve planning. Second, we shall demonstrate that by embedding a VIN within an image-processing network, challenging structured-prediction tasks from real images can be tackled successfully.

\vspace{-5pt}
\subsection{Mars Rover Navigation}\label{ss:mars}
\vspace{-5pt}
\input{mars_experiments}

\vspace{-0.5em}
\subsection{Continuous Control}\label{ss:continuous}
\vspace{-0.5em}
\input{continuous_experiments}

\vspace{-5pt}
\subsection{WebNav Challenge}\label{ss:webnav}
\vspace{-5pt}
\input{webnav_experiments}

%\section{Related Work}\label{s:related}

\vspace{-5pt}
\section{Conclusion and Outlook}\label{s:conclusion}
\vspace{-5pt}
\input{conclusion}
\subsubsection*{Acknowledgments}
This research was funded in part by Siemens, by ONR through a PECASE
award, by the Army Research
Office through the MAST program, and by an NSF CAREER award (\#1351028). A. T. was partially funded by the Viterbi Scholarship, Technion. Y. W. was partially funded by a DARPA PPAML program, contract FA8750-14-C-0011.

\newpage
{
\small
\setlength{\bibsep}{0.5pt plus 0.0ex}
\bibliography{valIter_nips16}
\bibliographystyle{plain}
}

\appendix
\input{appendix}

\end{document}

%% file: defs.tex
\newcommand{\cS}{\mathcal{S}}
\newcommand{\cA}{\mathcal{A}}

\newcommand{\ExpBy}[3]{\mathbb{E}^{#3}\left[ \left. #1 \right| #2\right]}

\DeclareMathOperator*{\argmax}{arg\,max}

%% file: intro.tex
% CNNs have been successful at visual tasks
Over the last decade, deep convolutional neural networks (CNNs) have revolutionized supervised learning for tasks such as object recognition, action recognition, and semantic segmentation \citep{Ciresan:2012b,krizhevsky2012imagenet,farabet2013learning,long2015fully}.
% Recently, CNNs have been applied to RL/control, with considerable success (DQN, GPS, TRPO).
Recently, CNNs have been applied to reinforcement learning (RL) tasks with visual observations such as Atari games \citep{mnih2015human}, robotic manipulation \citep{finn2016endtoend}, and imitation learning (IL) \citep{giusti2016machine}. In these tasks, a neural network (NN) is trained to represent a \emph{policy} -- a mapping from an observation of the system's state to an action, with the goal of representing a control strategy that has good \emph{long-term} behavior, typically quantified as the minimization of a sequence of time-dependent costs.

The \emph{sequential} nature of decision making in RL is inherently different than the one-step decisions in supervised learning, and in general requires some form of \emph{planning} \citep{Ber2012DynamicProgramming}. However, most recent deep RL works \citep{mnih2015human,finn2016endtoend,giusti2016machine} employed NN architectures that are very similar to the standard networks used in supervised learning tasks, which typically consist of CNNs for feature extraction, and fully connected layers that map the features to a probability distribution over actions. %For a control policy, the class labels are simply replaced by actions. 
Such networks are inherently \emph{reactive}, and in particular, lack explicit \emph{planning computation}. The success of reactive policies in sequential problems is due to the \emph{learning algorithm}, which essentially trains a reactive policy to select actions that have good long-term consequences in its training domain.

\begin{wrapfigure}{r}{0.5\textwidth}
\vspace{-5pt}
  \begin{center}
\includegraphics[scale=0.252,clip=true,trim=50 40 30 0]{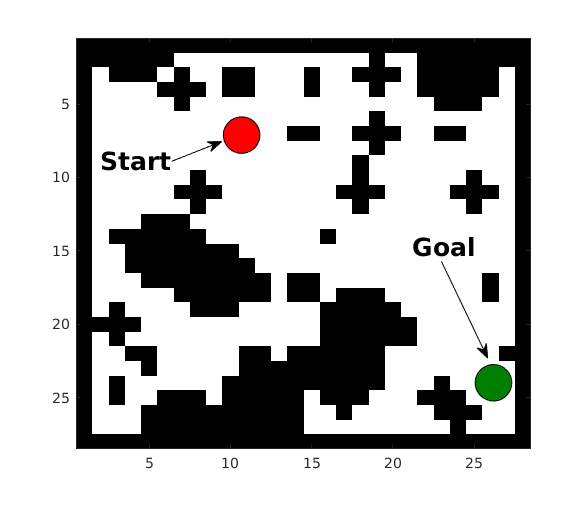}
\includegraphics[scale=0.272,clip=true,trim=50 40 30 0]{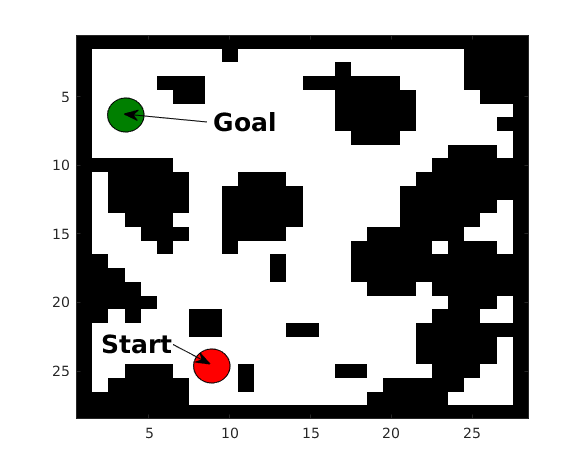}
  \end{center}
\vspace{-10pt}
  \caption{Two instances of a grid-world domain. Task is to move to the goal between the obstacles. \label{fig:fig_intro}   \vspace{-2em}
  }
\end{wrapfigure}

%%PA: I am not so convinced by the paragraph below; maybe cut it; maybe adjust
%%PA: first: hard to follow: it starts by saing 20 and 23 could solve this type of problems; but then later it says they can't (generalizes) -- but that's part of solving such problems?
%%PA: I don't see how the observation of them not solving these problems implies that the computation learned by reactive policies is different from planning?  I could see how one could directly claim that reactive policies don't plan, but I don't understand the reasoning that's being expressed here
%%PA: I think it'd be easier to argue for building in some prior knowledge about what good policies do in environments where lookahead can be helpful; and then say that VINs provide a suitable prior -- empiricially resulting in faster learning, but also not too strong (such that it's still flexible to learn great policies); then can also incorporate the footnote stuff in this paragraph; but not all of it; the only argument used could be: requires significantly less data (say something more specific based on our experimental findings)
To understand why planning can nevertheless be an important ingredient in a policy, consider the grid-world navigation task depicted in Figure \ref{fig:fig_intro} (left), in which the agent can observe a map of its domain, and is required to navigate between some obstacles to a target position. 
%Assuming that the policy can observe the domain, approaches such as in \citep{mnih2015human,finn2016endtoend} can learn a reactive policy that solves this task. 
One hopes that after training a policy to solve several instances of this problem with different obstacle configurations, the policy would generalize to solve a different, unseen domain, as in Figure \ref{fig:fig_intro} (right). 
However, as we show in our experiments, while standard CNN-based networks can be easily trained to solve a set of such maps, they 
do not generalize well to new tasks outside this set, because they do not understand the goal-directed nature of the behavior. 
%%SL.5.18: perhaps "do not succeed in this task" is too strong a term here?
This observation suggests that the computation learned by reactive policies is different from planning, which is required to solve a new task\footnote{In principle, with enough training data that covers all possible task configurations, and a rich enough policy representation, a reactive policy can learn to map each task to its optimal policy. In practice, this is often too expensive, and we offer a more data-efficient approach by exploiting a flexible prior about the planning computation underlying the behavior.}.

%The straightforward explanation for this is that the computation required for planning a path is very different from the reactive type of computation performed by these networks.

In this work, we propose a NN-based policy that can effectively \emph{learn to plan}. Our model, termed a \emph{value-iteration network} (VIN), has a differentiable `planning program' embedded within the NN structure.
%, which allows the network to learn an approximation of the planning computation relevant for the task.

The key to our approach is an observation that the classic value-iteration (VI)  
planning algorithm \citep{bellman1957dynamic,Ber2012DynamicProgramming} may be represented by a specific type of CNN. By embedding such a VI network module inside a standard feed-forward classification network, we obtain a NN model that can learn the parameters of a planning computation that yields useful predictions. The VI block is differentiable, and the whole network can be trained using standard backpropagation. This makes our policy simple to train using standard RL and IL algorithms, and straightforward to integrate with NNs for perception and control. 
%%SL.5.20: mention webnav here, I think that's quite important

%We emphasize that our contribution is a new \emph{policy representation} and not a new learning algorithm. Indeed, we show that VIN policies can be trained using standard model-free RL or imitation learning algorithms as in \citep{mnih2015human,finn2016endtoend}. After training, the network learns to map an observation to an underlying planning computation, and generate action predictions based on the resulting plan. As we demonstrate, this leads to policies that \emph{generalize better} across tasks. 
%In addition, we investigate a hierarchical extension of the VI block, which learns more efficiently on larger domains.

%\textbf{TBD: modify this paragraph with current results}. We show that value-iteration networks can learn non-trivial policies from images of synthetic grid-world domains, and directly from raw overhead images of Mars terrain, as in Figure \ref{fig:mars_intro}. Our approach performs significantly better than state-of-the-art reactive networks, and does not require knowing the planning model in advance. 

% An alternative approach is model based. This by definition generalizes to new problems, but requires SYSID, and solving the model. For many interesting problems such as robotic manipulation/locomotion this is challenging, motivating the success of model free methods such GPS/TRPO. Our method is essentially model free (although some knowledge can be used to pretrain the network) and doesn't require SYSID. It allows to potentially enjoy both worlds. 

Connections between planning algorithms and recurrent NNs were previously explored by Ilin et al.~\citep{ilin2007efficient}. Our work builds on related ideas, but results in a more broadly applicable policy representation. Our approach is different from model-based RL \citep{schmidhuber1990line,deisenroth2011pilco}, which requires system identification to map the observations to a dynamics model, which is then solved for a policy. In many applications, including robotic manipulation and locomotion, accurate system identification is difficult, and modelling errors can severely degrade the policy performance. 
%In addition, in most real-world problems the model can only be solved approximately, leading to further accuracy issues. 
In such domains, a model-free approach is often preferred \citep{finn2016endtoend}. 
Since a VIN is just a NN policy, it can be trained model free, without requiring explicit system identification. In addition, the effects of modelling errors in VINs can be mitigated by training the network end-to-end, similarly to the methods in \cite{joseph2013reinforcement,guo2016deep}. 
%We demonstrate this idea on a continuous path planning problem, in which trajectory planning is performed on a discrete 2-dimensional grid, and this plan is combined with a continuous NN actuation policy for moving the agent. 

We demonstrate the effectiveness of VINs within standard RL and IL algorithms in various problems, among which require visual perception, continuous control, and also natural language based decision making in the WebNav challenge~\citep{nogueira2016webnav}. After training, the policy learns to map an observation to a planning computation relevant for the task, and generate action predictions based on the resulting plan. As we demonstrate, this leads to policies that \emph{generalize better} to new, unseen, task instances. 

%% file: background.tex
In this section we provide background on planning, value iteration, CNNs, and policy representations for RL and IL. In the sequel, we shall show that CNNs can implement a particular form of planning computation similar to the value iteration algorithm, which can then be used as a policy for RL or IL.

\textbf{Value Iteration:}\label{ss:VI_background}
A standard model for sequential decision making and planning is the Markov decision process (MDP) \citep{bellman1957dynamic,Ber2012DynamicProgramming}.
 An MDP $M$ consists of states $s \in \cS$, actions $a\in \cA$, a reward function $R(s,a)$, and a transition kernel $P(s'|s,a)$ that encodes the probability of the next state given the current state and action. A policy $\pi(a|s)$ prescribes an action distribution for each state. 
The goal in an MDP is to find a policy that obtains high rewards in the \emph{long term}. Formally, the \emph{value} $V^\pi(s)$ of a state under policy $\pi$ is the expected discounted sum of rewards when starting from that state and executing policy $\pi$, 
$\small{V^\pi(s) \doteq \ExpBy{\sum_{t=0}^\infty \gamma^t r(s_t,a_t)}{s_0=s}{\pi}},$ where $\gamma \in (0,1)$ is a discount factor, and $\mathbb E^{\pi}$ denotes an expectation over trajectories of states and actions $(s_0,a_0,s_1,a_1\dots)$, in which actions are selected according to $\pi$, and states evolve according to the transition kernel $P(s'|s,a)$. The optimal value function \mbox{$V^*(s) \doteq \max_{\pi} V^\pi(s)$} is the maximal long-term return possible from a state. A policy $\pi^*$ is said to be optimal if  \mbox{$V^{\pi^*}(s) = V^*(s) \ \ \forall s$}. A popular algorithm for calculating $V^*$ and $\pi^*$ is value iteration (VI):
\vspace{-0.1em}
\begin{equation}\label{eq:VI_alg}
%\begin{split}
	\textstyle{V_{n+1}(s) = \max_a Q_n(s,a)  \quad \forall s,\quad \textrm{where} \quad
	Q_n(s,a) = R(s,a) + \gamma \sum_{s'} P(s'|s,a) V_n(s').}
%\end{split}
\vspace{-0.1em}
\end{equation}
It is well known that the value function $V_n$ in VI converges as $n\to \infty$ to $V^*$, from which an optimal policy may be derived as $\pi^*(s) = \argmax_a Q_\infty(s,a)$. 
%In practice, VI can only be applied for a finite number of iterations, yielding an approximate solution, for which error bounds are known \citep{Ber2012DynamicProgramming}.

\textbf{Convolutional Neural Networks (CNNs)}\label{ss:CNN_background}
are NNs with a particular architecture that has proved useful for computer vision, among other domains \citep{fukushima1979neural,lecun1998gradient,Ciresan:2012b,krizhevsky2012imagenet}. A CNN is comprised of stacked convolution and max-pooling layers. The input to each convolution layer is a 3-dimensional signal $X$, typically, an image with $l$ channels, $m$ horizontal pixels, and $n$ vertical pixels, and its output $h$ is a $l'$-channel convolution of the image with kernels $W^1,\dots,W^{l'}$,
%\begin{equation}\label{eq:conv_layer}
$\small{
	h_{l',i',j'} = \sigma \left( \sum_{l,i,j} W^{l'}_{l,i,j} X_{l,i'-i,j'-j} \right),}
$
%\end{equation}
where $\sigma$ is some scalar activation function. %, typically a rectified linear unit (ReLU): $\sigma(x)=\max(x,0)$.
A max-pooling layer selects, for each channel $l$ and pixel $i,j$ in $h$, the maximum value among its neighbors $N(i,j)$,
%\begin{equation}\label{eq:maxpool_layer}
$
	h^{maxpool}_{l,i,j} = \max_{i',j'\in N(i,j)} h_{l,i',j'}. 
$
%\end{equation}
Typically, the neighbors $N(i,j)$ are chosen as a $k \times k$ image patch around pixel $i,j$.
After max-pooling, the image is down-sampled by a constant factor $d$, commonly 2 or 4, resulting in an output signal with $l'$ channels, $m/d$ horizontal pixels, and $n/d$ vertical pixels.
CNNs are typically trained using stochastic gradient descent (SGD), with backpropagation for computing gradients.
\vspace{-2pt}

\textbf{Reinforcement Learning and Imitation Learning:}\label{ss:RL_background}
In MDPs where the state space is very large or continuous, or when the MDP transitions or rewards are not known in advance, planning algorithms cannot be applied. In these cases, a policy can be \emph{learned} from either expert supervision -- IL, or by trial and error -- RL. 
While the learning algorithms in both cases are different, the policy representations -- which are the focus of this work -- are similar. 
Additionally, most state-of-the-art algorithms such as \citep{ross2011reduction,mnih2015human,schulman2015trust,finn2016endtoend} are agnostic to the policy representation, and only require it to be differentiable, for performing gradient descent on some algorithm-specific loss function. Therefore, in this paper we do not commit to a specific learning algorithm, and only consider the policy.
\vspace{-2pt}

Let $\phi(s)$ denote an observation for state $s$. The policy is specified as a parametrized function $\pi_\theta(a|\phi(s))$ mapping observations to a probability over actions, where $\theta$ are the policy parameters. For example, the policy could be represented as a neural network, with $\theta$ denoting the network weights. The goal is to tune the parameters such that the policy behaves well in the sense that $\pi_\theta(a|\phi(s)) \approx \pi^*(a|\phi(s))$, where $\pi^*$ is the optimal policy for the MDP, as defined in Section \ref{ss:VI_background}.
\vspace{-2pt}
%\footnote{The popular Q-learning representation $\pi_\theta(a|\phi(s)) \doteq \argmax_a Q_\theta(\phi(s),a)$ \citep{mnih2015human,sutton1998reinforcement} where $Q_\theta(\phi(s),a)$ is a parametrized approximate $Q$ function}

In IL, a dataset of $N$ state observations and corresponding optimal actions $\small{\left\{\phi(s^i),a^i\sim \pi^*(\phi(s^i))\right\}_{i=1,\dots,N}}$ is generated by an expert. 
Learning a policy then becomes an instance of supervised learning \citep{ross2011reduction,giusti2016machine}.
In RL, the optimal action is not available, but instead, the agent can act in the world and observe the rewards and state transitions its actions effect. 
%$\left\{\phi(s^i),a^i,r^i,\phi(s^{i+1})\right\}_{i=1,\dots,N}$. 
RL algorithms such as in \citep{sutton1998reinforcement,mnih2015human,schulman2015trust,finn2016endtoend} use these observations to improve the value of the policy.
\vspace{-2pt}

%images $\left\{x^i\right\}_{i=1,\dots,N}$, states
%$\left\{s^i\right\}_{i=1,\dots,N}$, and action labels $\left\{ a^i \right\}_{i=1,\dots,N}$. We assume that the actions are generated from an optimal policy $a^i = \pi^*(s^i)$ with respect to some unknown MDP, encoded in the image $x^i$. For example, in the path planning domains we consider, the state $s$ encodes the position, and the image shows the obstacles and a goal location. The action then encodes the direction of the shortest path to the goal between the obstacles. We emphasize that the reward and transitions in the MDP are unknown, and are not an explicit part of the input in any way.

%Our goal is to learn a mapping $y(x,s)$ that best predicts the correct action label for an image and state. 
%\Naively, this problem can be cast as a standard supervised learning problem \cite{Bishop2006}, by concatenating $x$ and $s$ into a single observation. With enough training data, it is conceivable that a standard supervised learning algorithm would learn a suitable solution. However, such a \naive\ approach ignores the knowledge about the process that generates the data. By exploiting this knowledge, a solution may be found more efficiently. In the following, we propose a class of neural network representations that incorporate a `planning module' in their structure. We shall show that this structure enables much more efficient learning than standard feed-forward networks in this problem domain.

%% file: model.tex
In this section we introduce a general policy representation that embeds an explicit \emph{planning module}. As stated earlier, the motivation for such a representation is that a natural solution to many tasks, such as the path planning described above, involves planning on some model of the domain. 

%%SL.5.19: I'm a bit concerned that the phrasing here makes it unclear that \bar{M} is latent and that you don't have to specify it by hand. Maybe this would be clearer if you're more explicit about the underlying assumptions. For example, you could say that you assume there is some unknown underlying MDP \bar{M} such that, if you have a solution to \bar{M}, you have some information is very useful for choosing a good action in the original task. The trouble is that you don't know what \bar{M} is. However, you hypothesize that if you equip the policy with the ability to learn and solve \bar{M}, it will automatically learn the abstraction that embeds the space in \bar{M}. Making this assumption explicit would make the following paragraphs much easier to understand I think.
Let $M$ denote the MDP of the domain for which we design our policy $\pi$. We assume that there is some unknown MDP $\bar{M}$ such that the optimal plan in $\bar{M}$ contains useful information about the optimal policy in the original task $M$. However, we emphasize that we do not assume to know $\bar{M}$ in advance. Our idea is to equip the policy with the \emph{ability to learn and solve} $\bar{M}$, and to add the solution of $\bar{M}$ as an element in the policy $\pi$. We hypothesize that this will lead to a policy that automatically learns a useful $\bar{M}$ to plan on.
%abstraction that embeds the space in \bar{M}.
We denote by $\bar{s} \in \bar{S}$, $\bar{a} \in \bar{A}$, $\bar{R}(\bar{s},\bar{a})$, and $\bar{P}(\bar{s}'|\bar{s},\bar{a})$ the states, actions, rewards, and transitions in $\bar{M}$. To facilitate a connection between $M$ and $\bar{M}$, we let $\bar{R}$ and $\bar{P}$ depend on the observation in $M$, namely, $\bar{R} = f_R(\phi(s))$ and $\bar{P} = f_P(\phi(s))$, and we will later learn the functions $f_R$ and $f_P$ as a part of the policy learning process.

For example, in the grid-world domain described above, we can let $\bar{M}$ have the same state and action spaces as the true grid-world $M$. The reward function $f_R$ can map an image of the domain to a high reward at the goal, and negative reward near an obstacle, while $f_P$ can encode deterministic movements in the grid-world that do not depend on the observation. While these rewards and transitions are not necessarily the true rewards and transitions in the task, an optimal plan in $\bar{M}$ will still follow a trajectory that avoids obstacles and reaches the goal, similarly to the optimal plan in $M$. 

Once an MDP $\bar{M}$ has been specified, any standard planning algorithm can be used to obtain the value function $\bar{V}^*$. In the next section, we shall show that using a particular implementation of VI for planning has the advantage of being differentiable, and simple to implement within a NN framework. In this section however, we focus on how to use the planning result $\bar{V}^*$ within the NN policy $\pi$. Our approach is based on two important observations. The first is that the vector of values $\bar{V}^*(s) \ \ \forall s$ encodes all the information about the optimal plan in $\bar{M}$. Thus, adding the vector $\bar{V}^*$ as additional features to the policy $\pi$ is sufficient for extracting information about the optimal plan in $\bar{M}$. 

However, an additional property of $\bar{V}^*$ is that the optimal decision $\bar{\pi}^*(\bar{s})$ at a state $\bar{s}$ can depend only on a subset of the values of $\bar{V}^*$, since $\bar{\pi}^*(\bar{s}) = \argmax_{\bar{a}} \bar{R}(\bar{s},\bar{a}) + \gamma \sum_{\bar{s}'} \bar{P}(\bar{s}'|\bar{s},\bar{a}) \bar{V}^*(\bar{s}')$. Therefore, if the MDP has a local connectivity structure, such as in the grid-world example above, the states for which $\bar{P}(\bar{s}'|\bar{s},\bar{a})>0$ is a small subset of $\bar{S}$.

\begin{figure}[ht]
\vspace{-5pt}
\begin{center}
\includegraphics[scale=0.34,clip=true,trim=0 0 0 0]{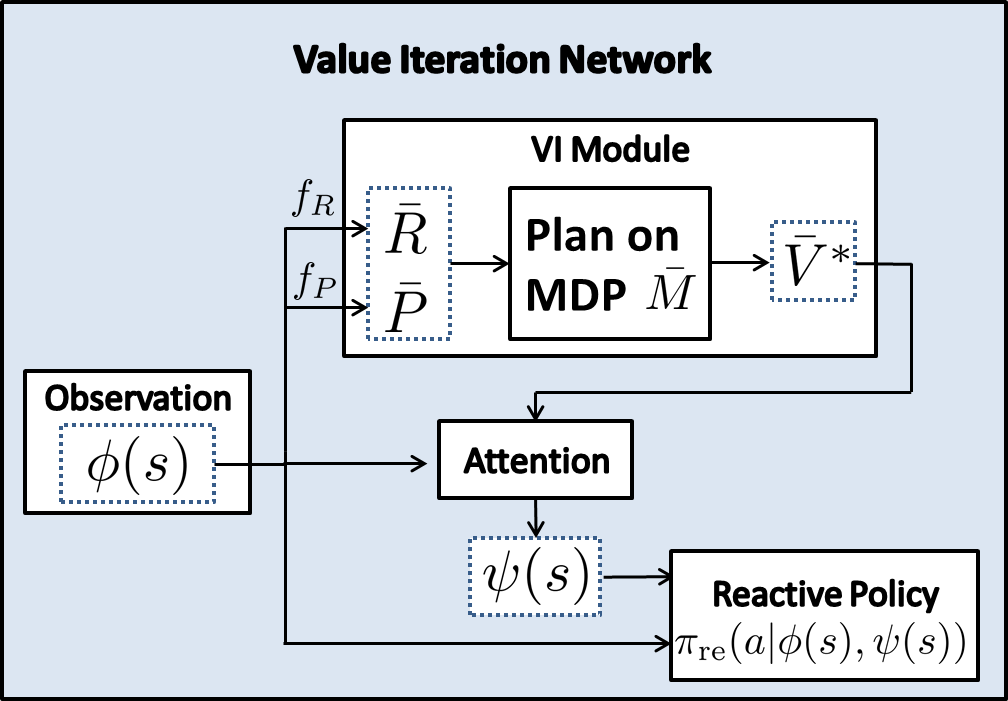}
\hspace{5pt}
\includegraphics[scale=0.63,clip=true,trim=180 342 255 50]{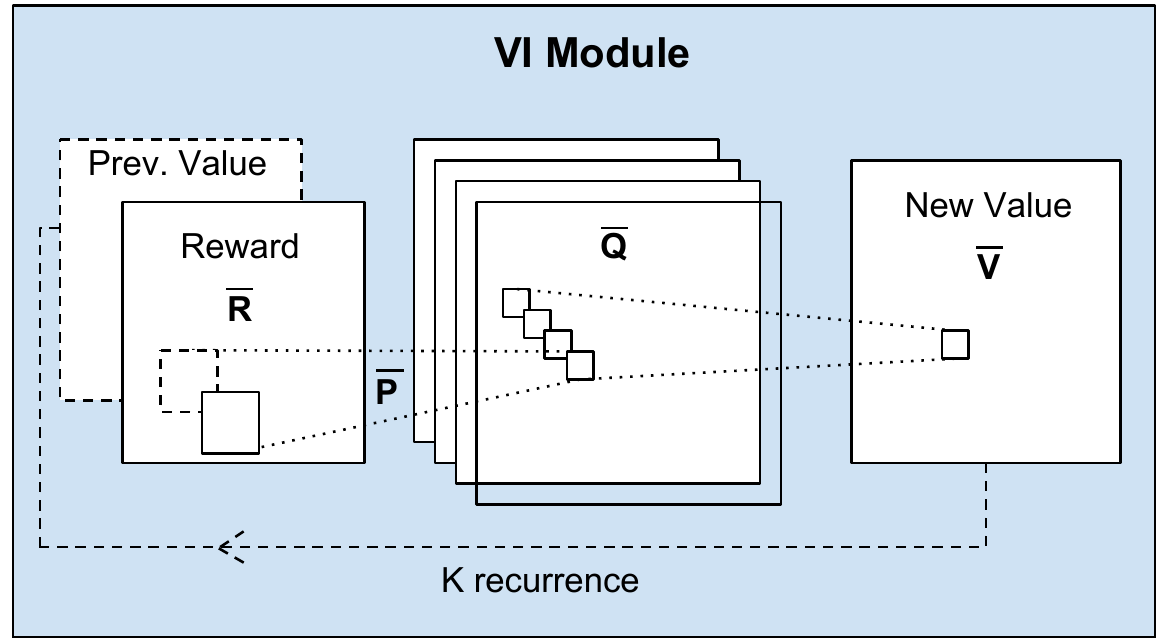}
\end{center}
\caption{Planning-based NN models. Left: a general policy representation that adds value function features from a planner to a reactive policy. Right: VI module -- a CNN representation of VI algorithm. \label{fig:vin_schematic}   }
\vspace{-15pt}
\end{figure}

In NN terminology, this is a form of \emph{attention} \citep{xu2015show}, in the sense that for a given label prediction (action), only a subset of the input features (value function) is relevant. Attention is known to improve learning performance by reducing the effective number of network parameters during learning. Therefore, the second element in our network is an attention module %$\Pi(\bar{V}^*,\phi(s))$, which
that outputs a vector of (attention modulated) values $\psi(s)$. Finally, the vector $\psi(s)$ is added as additional features to a reactive policy $\pi_{\text{re}}(a|\phi(s),\psi(s))$. The full network architecture is depicted in Figure \ref{fig:vin_schematic} (left).

Returning to our grid-world example, at a particular state $s$, the reactive policy only needs to query the values of the states neighboring $s$ in order to select the correct action. Thus, the attention module in this case could return a $\psi(s)$ vector with a subset of $\bar{V}^*$ for these neighboring states.

%%PA: why are some lines thick, some thin?

Let $\theta$ denote all the parameters of the policy, namely, the parameters of $f_R$, $f_P$, %$\Pi$, 
and $\pi_{\text{re}}$, and note that $\psi(s)$ is in fact a function of $\phi(s)$. Therefore, the policy can be written in the form $\pi_\theta(a|\phi(s))$, similarly to the standard policy form (cf. Section \ref{ss:RL_background}). If we could back-propagate through this function, then potentially we could train the policy using standard RL and IL algorithms, just like any other standard policy representation. While it is easy to design functions $f_R$ and $f_P$
%, and $\Pi$ 
that are differentiable (and we provide several examples in our experiments), back-propagating the gradient through the planning algorithm is not trivial. In the following, we propose a novel interpretation of an approximate VI algorithm as a particular form of a CNN. This allows us to conveniently treat the planning module as just another NN, and by back-propagating through it, we can train the whole policy \emph{end-to-end}.

\vspace{-10pt}
\subsection{The VI Module}\label{ss:VI_module}
\vspace{-5pt}
We now introduce the VI module -- a NN that encodes a differentiable planning computation. 
\vspace{-5pt}

Our starting point is the VI algorithm \eqref{eq:VI_alg}. Our main observation is that each iteration of VI may be seen as passing the previous value function $V_n$ and reward function $R$ through a convolution layer and max-pooling layer. In this analogy, each channel in the convolution layer corresponds to the $Q$-function for a specific action, and convolution kernel weights correspond to the discounted transition probabilities. Thus by recurrently applying a convolution layer $K$ times, $K$ iterations of VI are effectively performed.

Following this idea, we propose the VI network module, as depicted in Figure \ref{fig:vin_schematic}B. The inputs to the VI module is a `reward image' $\bar{R}$ of dimensions $l,m,n$, where 
%and a state $(i_s, j_s) \in (1,\dots,m)\times(1,\dots,n)$. 
here, for the purpose of clarity, we follow the CNN formulation and explicitly assume that the state space $\bar{S}$ maps to a 2-dimensional grid. However, our approach can be extended to general discrete state spaces, for example, a graph, as we report in the WikiNav experiment in Section \ref{ss:webnav}. 
The reward is fed into a convolutional layer $\bar{Q}$ with $\bar{A}$ channels and a linear activation function,
%\begin{equation*}
$
	\bar{Q}_{\bar{a},i',j'} = \sum_{l,i,j} W^{\bar{a}}_{l,i,j} \bar{R}_{l,i'-i,j'-j}.
$
%\end{equation*}
Each channel in this layer corresponds to $\bar{Q}(\bar{s},\bar{a})$ for a particular action $\bar{a}$. This layer is then max-pooled along the actions channel to produce the next-iteration value function layer $\bar{V}$,
%\begin{equation*}
$
	\bar{V}_{i,j} = \max_{\bar{a}} \bar{Q}(\bar{a},i,j).
$
%\end{equation*}
The next-iteration value function layer $\bar{V}$ is then stacked with the reward $\bar{R}$, and \emph{fed back} into the convolutional layer and max-pooling layer $K$ times, to perform $K$ iterations of value iteration. 

%After $K$ such recurrences, the $A$ channels of the $q$ layer in the $i_s,j_s$ position are fed into a fully connected softmax output layer $y$,
%\begin{equation*}
%	y(a) = \frac{\exp\left( \sum_{a'} W^{out}_{a',a} q(a',i_s,j_s) \right)}{\sum_{a} \exp\left(\sum_{a'} W^{out}_{a',a} q(a',i_s,j_s) \right)}.
%\end{equation*}

%\begin{figure}[ht]
%\begin{center}
%\includegraphics[scale=0.53,clip=true,trim=180 260 %160 50]{VI_block}
%\end{center}
%\caption{The VI block. A neural-network that approximates the VI algorithm. \label{fig:vi_block}   }
%\end{figure}

The VI module is simply a NN architecture that has the capability of performing an approximate VI computation. Nevertheless, representing VI in this form makes \emph{learning} the MDP parameters and reward function natural -- by backpropagating through the network, similarly to a standard CNN. 
VI modules can also be composed hierarchically, by treating the value of one VI module as additional input to another VI module. We further report on this idea in the supplementary material.

\vspace{-5pt}
\subsection{Value Iteration Networks}
\vspace{-5pt}
We now have all the ingredients for a differentiable planning-based policy, which we term a value iteration network (VIN). The VIN is based on the general planning-based policy defined above, with the VI module as the planning algorithm. 
In order to implement a VIN, one has to specify the state and action spaces for the planning module $\bar{S}$ and $\bar{A}$, the reward and transition functions $f_R$ and $f_P$, and the attention function; we refer to this as the \emph{VIN design}. For some tasks, as we show in our experiments, it is relatively straightforward to select a suitable design, while other tasks may require more thought. However, we emphasize an important point: the reward, transitions, and attention can be defined by parametric functions, and trained with the whole policy\footnote{VINs are fundamentally different than inverse RL methods~\cite{neu2007apprenticeship}, where transitions are required to be known.}. Thus, a rough design can be specified, and then fine-tuned by end-to-end training.

Once a VIN design is chosen, implementing the VIN is straightforward, as it is simply a form of a CNN. The networks in our experiments all required only several lines of Theano \cite{Bastien-Theano-2012} code. 
%%GT: perhaps "a few" instead of "several", or perhaps "a relatively small number"? The phrase "only several" sounds off to me
In the next section, we evaluate VIN policies on various domains, showing that by learning to plan, they achieve a better generalization capability.

%% file: gridworld_experiments.tex
\begin{figure}
\begin{center}
\includegraphics[scale=0.3,clip=true,trim=55 40 42 29]{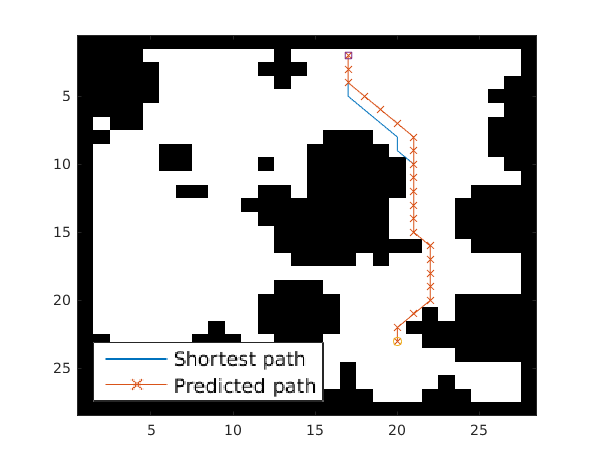}
\hspace{5pt}
\includegraphics[scale=0.315, clip=true,trim=65 40 58 40]{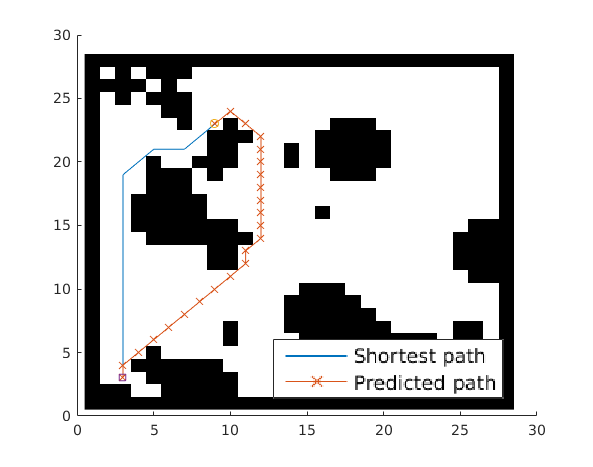}
\hspace{5pt}
\includegraphics[scale=0.465,clip=true,trim=130 115 135 10]{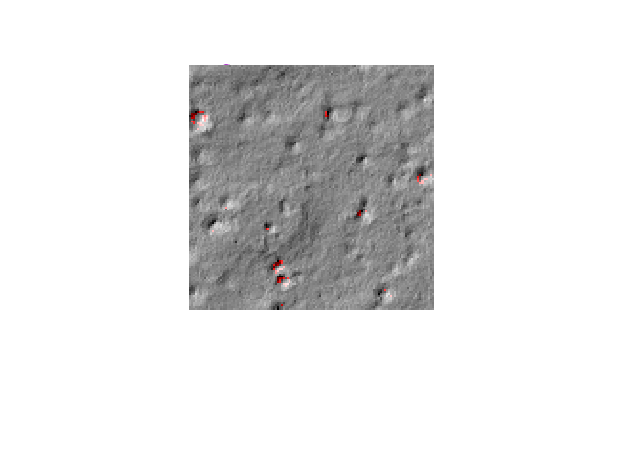}
\hspace{2pt}
\includegraphics[scale=0.465,clip=true,trim=130 115 135 10]{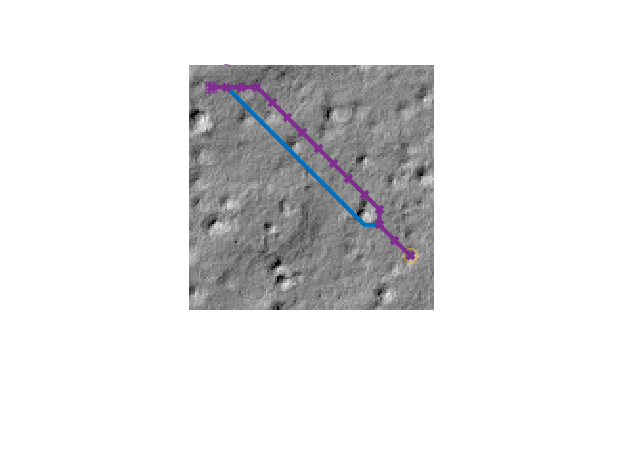}
\end{center}
\vspace{-10pt}
\caption{Grid-world domains (best viewed in color). A,B: Two random instances of the $28 \times 28$ synthetic gridworld, with the VIN-predicted trajectories and ground-truth shortest paths between random start and goal positions. %In A, the trajectories mostly overlap. Note that the shortest-path trajectory is not unique, and in the left domain the trajectory predicted by the VIN is different than the ground-truth, but of the same length. 
C: An image of the Mars domain, with points of elevation sharper than $10^{\circ}$ colored in red. These points were calculated from a matching image of elevation data (not shown), and were not available to the learning algorithm. Note the difficulty of distinguishing between obstacles and non-obstacles. D: The VIN-predicted (purple line with cross markers), and the shortest-path ground truth  (blue line) trajectories between between random start and goal positions. \label{fig:gridworld}   }
\vspace{-10pt}
\end{figure}

\setlength{\parskip}{3pt}
Our first experiment domain is a synthetic grid-world with randomly placed obstacles, in which the observation includes the position of the agent, and also an image of the map of obstacles and goal position. Figure \ref{fig:gridworld}
shows two random instances of such a grid-world of size $16 \times 16$. We conjecture that by learning the optimal policy for several instances of this domain, a VIN policy would learn the planning computation required to solve a new, unseen, task.

In such a simple domain, an optimal policy can easily be calculated using exact VI. Note, however, that here we are interested in evaluating whether a NN policy, trained using RL or IL, can \emph{learn to plan}. In the following results, policies were trained using IL, by standard supervised learning from demonstrations of the optimal policy. In the supplementary material, we report additional RL experiments that show similar findings.
%To cleanly evaluate this, we chose to train the policy using IL, by standard supervised learning from demonstrations of the optimal policy. In the supplementary material, we provide additional RL experiments that show similar findings.

We design a VIN for this task following the guidelines described above, where the planning MDP $\bar{M}$ is a grid-world, similar to the true MDP. The reward mapping $f_R$ is a CNN mapping the image input to a reward map in the grid-world. Thus, $f_R$ should potentially learn to discriminate between obstacles, non-obstacles and the goal, and assign a suitable reward to each.
The transitions $\bar{P}$ were defined as $3 \times 3$ convolution kernels in the VI block, exploiting the fact that transitions in the grid-world are local\footnote{Note that the transitions defined this way do not depend on the state $\bar{s}$. Interestingly, we shall see that the network learned to plan successful trajectories nevertheless, by appropriately shaping the reward.}. The recurrence $K$ was chosen in proportion to the grid-world size, to ensure that information can flow from the goal state to any other state. For the attention module, we chose a trivial approach that selects the $\bar{Q}$ values in the VI block for the current state, i.e., $\psi(s) = \bar{Q}(s,\cdot)$. The final reactive policy is a fully connected network that maps $\psi(s)$ to a probability over actions.

%In this simple domain, the state space for the planning module $\bar{S}$ is equivalent to the real state space $S$, and the attention module is trivial. In addition, an optimal policy can easily be calculated, and used as an expert for supervised (imitation) learning. This allows us to cleanly investigate the learning capability of VINs, and to compare them with other state-of-the-art policy representations.

%We consider a $m \times n$ grid-world with randomly placed obstacles, and a border of size $1$. Figure \ref{fig:gridworld}
%shows an instance of the grid world with $m=n=16$. The possible actions are moving in one of the 8 directions. Attempting to move into an obstacle results in no movement. Finding a shortest path between two locations in the grid-world may be done using value iteration (among other standard planning algorithms), by using a constant negative reward for every non-goal state, and some positive reward for the goal.

%Our goal is to learn a policy that maps from an image of the grid-world (as in Figure \ref{fig:gridworld}), the current position in the map, and some goal position, to an action that leads to the shortest path to the goal. We use supervised learning for this task. 

We compare VINs to the following NN reactive policies:

\textbf{CNN network:}
We devised a CNN-based reactive policy inspired by the recent impressive results of DQN \cite{mnih2015human}, with 5 convolution layers, and a fully connected output. While the network in \cite{mnih2015human} was trained to predict Q values, our network outputs a probability over actions. These terms are related, since $\pi^*(s) = \argmax_a Q(s,a)$.
\textbf{Fully Convolutional Network (FCN):}
The problem setting for this domain is similar to semantic segmentation \citep{long2015fully}, in which each pixel in the image is assigned a semantic label (the action in our case). We therefore devised an FCN inspired by a state-of-the-art semantic segmentation algorithm \citep{long2015fully}, with 3 convolution layers, where the first layer has a filter that spans the whole image, to properly convey information from the goal to every other state.

In Table \ref{tab:gridworld} we present the average $0-1$ prediction loss of each model, evaluated on a held-out test-set of maps with random obstacles, goals, and initial states, for different problem sizes. 
In addition, for each map, a full trajectory from the initial state was predicted, by iteratively rolling-out the next-states predicted by the network. A trajectory was said to succeed if it reached the goal without hitting  obstacles. For each trajectory that succeeded, we also measured its difference in length from the optimal trajectory. The average difference and the average success rate are reported in Table \ref{tab:gridworld}.

Clearly, VIN policies generalize to domains outside the training set. A visualization of the reward mapping $f_R$ (see supplementary material) shows that it is negative at obstacles, positive at the goal, and a small negative constant otherwise. The resulting value function has a gradient pointing towards a direction to the goal around obstacles, thus a useful planning computation was learned.
VINs also significantly outperform the reactive networks, and the performance gap increases dramatically with the problem size. Importantly, note that the prediction loss for the reactive policies is comparable to the VINs, although their success rate is significantly worse. This shows that this is not a standard case of overfitting/underfitting of the reactive policies. Rather, VIN policies, by their VI structure, focus prediction errors on less important parts of the trajectory, while reactive policies do not make this distinction, and learn the easily predictable parts of the trajectory yet fail on the complete task.

The VINs have an effective depth of $K$, which is larger than the depth of the reactive policies. One may wonder, whether any deep enough network would learn to plan. In principle, a CNN or FCN of depth $K$ has the potential to perform the same computation as a VIN. However, it has much more parameters, requiring much more training data. We evaluate this by untying the weights in the $K$ recurrent layers in the VIN. Our results, reported in the supplementary material, show that untying the weights degrades performance, with a stronger effect for smaller sizes of training data.

%we report the actual performance in predicting a trajectory by measuring the average success rate in reaching the goal, and the length of the trajectory compared to the optimal shortest-path. 
%We evaluated these measures on a held-out test-set of 1000 maps with random obstacles, goal positions, and initial states. 

%The $8 \times 8$ domain is small enough for a reactive policy to `remember' all possible obstacle configurations, and perform reasonably well. The number of possible configurations, however, increases exponentially with the problem size, leading to a worse performance on the larger maps. The VIN, on the other hand, overcomes this issue by exploiting the \emph{structure} of the problem.

%We emphasize that the model (i.e., transition dynamics) used for planning the shortest-path trajectories was not given to the learning algorithm at any point. %In such a setting, exact IRL is impossible.

\begin{table}
%\vspace{-10pt}
\begin{center}
  \begin{tabular}{ | c | c | c | c | c | c | c | c | c | c |}
    \hline
    \multirow{2}{*}{Domain} 
    & \multicolumn{3}{|c|}{VIN} 
    & \multicolumn{3}{|c|}{CNN} 
    & \multicolumn{3}{|c|}{FCN} \\ \cline{2-10}
     & Prediction & Success & Traj. 
     & Pred. & Succ. & Traj. 
     & Pred. & Succ. & Traj. \\ 
     & loss & rate & diff. 
     & loss & rate & diff. 
     & loss & rate & diff. \\ \hline 

    $8 \times 8$ & 0.004 & \bf 99.6\% & 0.001 
    & 0.02 & 97.9\% & 0.006 
    & 0.01 & 97.3\%& 0.004\\ \hline 
    $16 \times 16$ & 0.05 & \bf 99.3\% & 0.089 
    & 0.10 & 87.6\% & 0.06 
    & 0.07 & 88.3\% & 0.05 \\ \hline 
    $28 \times 28$ & 0.11 & \bf 97\% & 0.086 
    & 0.13 & 74.2\% & 0.078 
    & 0.09 & 76.6\% & 0.08 \\ \hline 
  \end{tabular}
\end{center}
\caption{Performance on grid-world domain. Top: comparison with reactive policies. For all domain sizes, VIN networks significantly outperform standard reactive networks. Note that the performance gap increases dramatically with problem size. 
\vspace{-20pt}
%Bottom: evaluation of the effect of VI module shared weights relative to data size.
\label{tab:gridworld}   }
\end{table}

%% file: mars_experiments.tex
%%PA: weird to have a 4.1.1 but no 4.1.2 -- aren't their two sets of experiments under 4.1? with artifical gridworld and with mars domain?  maybe have two subsections accordingly?
% The image input to the previous grid-world experiments was a perfect and synthetic map of the obstacles. 
In this experiment we show that VINs can learn to plan \emph{from natural image input}. We demonstrate this on path-planning from overhead terrain images of a Mars landscape.

Each domain is represented by a $128\times 128$ image patch, on which we defined a $16 \times 16$ grid-world, where each state was considered an obstacle if the terrain in its corresponding $8\times 8$ image patch contained an elevation angle of $10$ degrees or more, evaluated using an external elevation data base. An example of the domain and terrain image is depicted in Figure \ref{fig:gridworld}. The MDP for shortest-path planning in this case is similar to the grid-world domain of Section \ref{ss:grid-world}, and the VIN design was similar, only with a deeper CNN in the reward mapping $f_R$ for processing the image. 

The policy was trained to predict the shortest-path \emph{directly from the terrain image}. We emphasize that the elevation data \emph{is not part of the input}, and 
%the elevation therefore 
must be inferred (if needed) from the terrain image.
%itself.

%We report our results in Table \ref{tab:mars} and Figure \ref{fig:gridworld}. 
After training, VIN achieved a success rate of $84.8$\%. To put this rate in context, we compare with the best performance achievable without access to the elevation data, which is $90.3$\%. To make this comparison, we trained a CNN to classify whether an $8 \times 8$ patch is an obstacle or not. This classifier was trained using the same image data as the VIN network, but its labels were the true obstacle classifications from the elevation map (we reiterate that the VIN \emph{did not} have access to these ground-truth obstacle labels during training or testing). %Training this classifier is a standard binary classification problem, and its performance represents the best obstacle identification possible with our CNN in this domain. 
The success rate of planner that uses the obstacle map generated by this classifier from the raw image is $90.3$\%, %The results of this optimal predictor are reported in Table \ref{tab:gridworld}. The 90.3\% success rate 
showing that obstacle identification from the raw image is indeed challenging. Thus, the success rate of the VIN, which was trained without any obstacle labels, and had to `figure out' the planning process is quite remarkable.

%\begin{figure*}
%\begin{center}
%\includegraphics[scale=0.47,clip=true,trim=130 100 130 10]{mars_3_obs_edit}
%\hspace{10pt}
%\includegraphics[scale=0.47,clip=true,trim=130 100 130 10]{mars_3_clean}
%\hspace{10pt}
%\includegraphics[scale=0.47,clip=true,trim=130 100 130 10]{mars_3_clan_traj}
%\end{center}
%\caption{Mars domain (best viewed in color). Left: a sample 128 by 128 image patch of the Mars terrain, with points of elevation sharper than $10^{\circ}$ colored in red. These points were calculated from a matching image of elevation data (not shown), and were not available to the learning algorithm. Middle: the actual image that was used for training the network. Note the difficulty of distinguishing between obstacles (red dots in the left figure) and non-obstacles. Right: the trajectory predicted by the trained VIN, between a random starting and goal positions (purple line, with cross markers), and the shortest-path ground truth trajectory between these points (blue line).
%\label{fig:mars_}   }
%\end{figure*}

%% file: continuous_experiments.tex
\begin{wrapfigure}{r}{.5\columnwidth}
\setlength{\unitlength}{0.5\columnwidth}
\centering
\vspace{-1em}
\begin{tabular}[c]{ | c | c | c | }
    \hline
    Network & Train Error & Test Error\\ \hline 

    VIN & 0.30 & 0.35 \\ \hline 
    CNN & 0.39 & 0.59  \\ \hline 
 \end{tabular}
\includegraphics[scale=0.267,clip=true,trim=30 0 50 0]{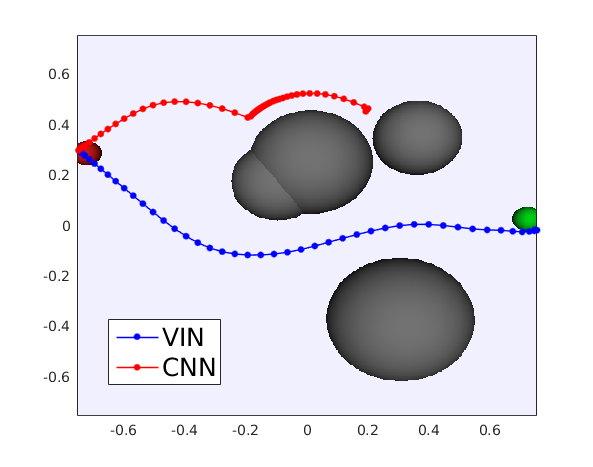}~
\includegraphics[scale=0.267,clip=true,trim=30 0 50 0]{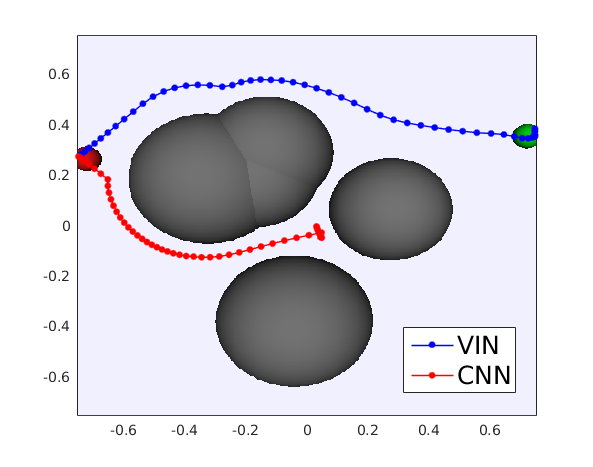}
\vspace{-1em}
\caption{Continuous control domain. Top: average distance to goal on training and test domains for VIN and CNN policies. Bottom: trajectories predicted by VIN and CNN on test domains.
\label{fig:continuous}   }
\vspace{-1em}
\end{wrapfigure}
% A simplifying factor in the previous experiments is that the planning was in a grid-world MDP similar to the true MDP. 
We now consider a 2D path planning domain with \emph{continuous states and continuous actions}, which cannot be solved using VI, and therefore a VIN cannot be naively applied. Instead, we will construct the VIN to perform `high-level' planning on a discrete, coarse, grid-world representation of the continuous domain. We shall show that a VIN can \emph{learn to plan} such a `high-level' plan, and also exploit that plan within its `low-level' continuous control policy. Moreover, the VIN policy results in better generalization than a reactive policy.

Consider the domain in Figure \ref{fig:continuous}. A red-colored particle needs to be navigated to a green goal using horizontal and vertical forces. Gray-colored obstacles are randomly positioned in the domain, and apply an elastic force and friction when contacted. This domain presents a non-trivial control problem, as the agent needs to both \emph{plan} a feasible trajectory between the obstacles (or use them to bounce off), but also control the particle (which has mass and inertia) to follow it. The state observation consists of the particle's continuous position and velocity, and a static $16 \times 16$ downscaled image of the obstacles and goal position in the domain. In principle, such an observation is sufficient to devise a `rough plan' for the particle to follow.

As in our previous experiments, we investigate whether a policy trained on several instances of this domain with different start state, goal, and obstacle positions, would generalize to an unseen domain.

For training we chose the guided policy search (GPS) algorithm with unknown dynamics \citep{levine2014learning}, which is suitable for learning policies for continuous dynamics with contacts, and we used the publicly available GPS code \citep{fzftm-gpsi-16}, and Mujoco \citep{todorov2012mujoco} for physical simulation. 
%GPS works by learning time-varying iLQG controllers for each domain, and then fitting the controllers to a single NN policy using supervised learning. This process is repeated for several iterations, and a special cost function is used to enforce an agreement between the trajectory distribution of the iLQG and NN controllers. We refer to \citep{levine2014learning,fzftm-gpsi-16} for the full algorithm details. For our task, we ran 10 iterations of iLQG, followed by one iteration of NN policy fitting. This allows us to cleanly compare VINs to other policies without GPS-specific effects. 
We generated 200 random training instances, 
%with randomly placed obstacles, and start and goal positions placed randomly on the borders of the domain. We 
and evaluate our performance on 40 \emph{different} test instances from the same distribution.

% \begin{figure}\centering
% \begin{minipage}{0.45\textwidth}
% \begin{tabular}[c]{ | c | c | c | }
%     \hline
%     Network & Train Error & Test Error\\ \hline 

%     VIN & 0.30 & 0.35 \\ \hline 
%     CNN & 0.19 & 0.73  \\ \hline 
%  \end{tabular}
% \end{minipage}
% \begin{minipage}{0.5\textwidth}
% \includegraphics[scale=0.267,clip=true,trim=30 0 50 0]{vin_211}
% %\includegraphics[scale=0.235,clip=true,trim=30 0 50 0]{dqn_211}
% \includegraphics[scale=0.267,clip=true,trim=30 0 50 0]{vin_225}
% %\includegraphics[scale=0.25,clip=true,trim=30 0 50 0]{dqn_225}
% %\end{center}
% \end{minipage}
% \vspace{-10pt}
% \caption{Continuous control domain. Left: average distance to goal on training and test domains for VIN and CNN policies. Right: trajectories predicted by VI and CNN on random test domains.
% \label{fig:continuous}   }
% \vspace{-20pt}
% \end{figure}

Our VIN design is similar to the grid-world cases, with some important modifications: the attention module selects a $5 \times 5$ patch of the value $\bar{V}$, centered around the current (discretized) position in the map. The final reactive policy is a 3-layer fully connected network, with a 2-dimensional continuous output for the controls. 
%The state space $\bar{S}$ is a $16 \times 16$ grid-world, and the transitions $\bar{P}$ are $3 \times 3$ convolution kernels in the VI block, similar to the grid-world of Section \ref{ss:grid-world}. In this case, 
In addition, due to the limited number of training domains, we pre-trained the VIN with transition weights that correspond to discounted grid-world transitions. 
%(for example, the transitions for an action to go north-west would be $\gamma$ in the top left corner and zeros otherwise), before training end-to-end. 
This is a reasonable prior for the weights in a 2-d task, and we emphasize that even with this initialization, the initial value function is meaningless, since the reward map $f_R$ is not yet learned.
%The reward mapping $f_R$ is a CNN with $s_{\text{image}}$ as its input, one layer with $150$ kernels of size $3 \times 3$, and a second layer with one $3 \times 3$ filter to output $\bar{R}$. 
We compare with a CNN-based reactive policy inspired by the state-of-the-art results in \citep{mnih2015human,mnih2016asynchronous}, with 2 CNN layers for image processing, followed by a 3-layer fully connected network similar to the VIN reactive policy.

Figure \ref{fig:continuous} shows the performance of the trained policies, measured as the final distance to the target. The VIN clearly outperforms the CNN on test domains.
% Note that while the CNN policy does well on the training domains, it is significantly outperformed by the VIN policy on the test domains. 
We also plot several trajectories of both policies on test domains, showing that VIN learned a more sensible generalization of the task. 
%This result demonstrates the importance of planning in generalizing to new tasks, and motivates the use of planning-based policies such as VINs. 
%%SL.5.20: I guess the moral of the story here is that the VIN incorporates more bias in favor of planning-style tasks, hence the worse performance on training domains and better generalization. One subtle but valuable point to make here is that in reinforcement learning and learning from demonstration, this is a *good* tradeoff to make, because data tends to be somewhat scarce.

%\begin{table}
%\begin{center}
%  \begin{tabular}{ | c | c | c | }
%    \hline
%    Network & Training Error & Test Error \\ 
%     &  & \\ \hline 
%    VIN & 0.30 & 0.35 \\ \hline 
%    DQN & 0.19 & 0.73  \\ \hline 
%  \end{tabular}
%\end{center}
%\caption{Performance on continuous control domain. \label{tab:continuous}   }
%\end{table}

%% file: webnav_experiments.tex
In the previous experiments, the planning aspect of the task corresponded to 2D navigation.
%%SL.5.20: be very careful here, this can be misread to say that the continuous domain is a gridworld, which is not true. Maybe rather say that: although the continuous task requires fine control that is not possible using only a discretization, the planning aspect of the task still corresponds to 2D navigation.
We now consider a more general domain: WebNav~\cite{nogueira2016webnav} -- a language based search task on a graph.

In WebNav~\cite{nogueira2016webnav}, the agent needs to navigate the links of a website towards a goal web-page, specified by a short 4-sentence query. At each state $s$ (web-page), the agent can observe average word-embedding features of the state $\phi(s)$ and possible next states $\phi(s')$ (linked pages), and the features of the query $\phi(q)$, and based on that has to select which link to follow. In \cite{nogueira2016webnav}, the search was performed on the Wikipedia website. Here, we report  experiments on the `Wikipedia for Schools' website, a simplified Wikipedia designed for children, with over 6000 pages and at most 292 links per page.

In \cite{nogueira2016webnav}, a NN-based policy was proposed, which first learns a NN mapping from $(\phi(s),\phi(q))$ to a hidden state vector $h$. The action is then selected according to $\pi(s'|\phi(s),\phi(q)) \propto \exp \left(h^\top \phi(s') \right)$. In essence, this policy is reactive, and relies on the word embedding features at each state to contain meaningful information about the path to the goal. Indeed, this property  naturally holds for an encyclopedic website that is structured as a tree of categories, sub-categories, sub-sub-categories, etc.

We sought to explore whether planning, based on a VIN, can lead to better performance in this task, with the intuition that a plan on a simplified model of the website can help guide the reactive policy in difficult queries. Therefore, we designed a VIN that plans on a small subset of the graph that contains only the 1st and 2nd level categories ($< 3$\% of the graph), and their word-embedding features. 
%The memory required for the VIN graph is comparable to the size of the baseline policy

Designing this VIN requires a different approach from the grid-world VINs described earlier, where the most challenging aspect is to define a meaningful mapping between nodes in the true graph and nodes in the smaller VIN graph. 
For the reward mapping $f_R$, we chose a weighted similarity measure between the query features $\phi(q)$, and the features of nodes in the small graph $\phi(\bar{s})$. Thus, intuitively, nodes that are similar to the query should have high reward. The transitions were fixed based on the graph connectivity of the smaller VIN graph, which is known, though different from the true graph.
%, and were not learned.
%%SL.5.20: again, be careful: this is not the known true graph, but a surrogate graph. Maybe make an analogy to semi-supervised learning here?
The attention module was also based on a weighted similarity measure between the features of the possible next states $\phi(s')$ and the features of each node in the simplified graph $\phi(\bar{s})$. The reactive policy part of the VIN was similar to the policy of \citep{nogueira2016webnav} described above.
Note that by training such a VIN end-to-end, we are effectively learning how to exploit the small graph for doing better planning on the true, large graph.
%The weights in the reward and attention mappings were learned end-to-end.
%%SL.5.20: this last sentence doesn't belong in this paragraph, it should go to where the results are -- this paragraph is for describing VIN

Both the VIN policy and the baseline reactive policy were trained by supervised learning, on random trajectories that start from the root node of the graph. Similarly to \citep{nogueira2016webnav}, a policy is said to succeed a query if all the correct predictions along the path are within its top-4 predictions. 

After training, the VIN policy performed mildly better than the baseline on 2000 held-out test queries when starting from the root node, achieving 1030 successful runs vs. 1025 for the baseline.
%%SL.5.20: are these "test" queries? if so, we should say that
However, when we tested the policies on a harder task of starting from a random position in the graph, VINs significantly outperformed the baseline, achieving 346 successful runs vs. 304 for the baseline, out of 4000 test queries. These results confirm that indeed, when navigating a tree of categories from the root up, the features at each state  contain meaningful information about the path to the goal, making a reactive policy sufficient. However, when starting the navigation from a different state, a reactive policy may fail to understand that it needs to first go back to the root and switch to a different branch in the tree. Our results indicate such a strategy can be better represented by a VIN.

We remark that there is still room for further improvements of the WebNav results, e.g., by better models for reward and attention functions, and better word-embedding representations of text.
%. Also, the word-embedding representations of text impact performance: we observed that just identifying whether a query is from a page or not is challenging.
%However, we believe that the most critical issue for WebNav is the word-embedding representation of the text: we observed that just identifying whether a query is from a page or not is challenging.
%%SL.5.20: if you have space, may be good here to play up the fact that this is a challenging real-world domain, and that incorporating a small amount of prior knowledge from the reduced size graph ("semi-supervised") is highly nontrivial using traditional methods?

%We remark that there is still large room for further improvements of the WebNav results, e.g., by better models for the reward and attention functions. However, we believe that the most critical issue for WebNav is the word-embedding representation of the text: we observed that just identifying whether a query is from a page or not is challenging.
%when only considering the bag-of-words model and the feature vector $\phi(\cdot)$, it remains very challenging to even distinguish when a query is from a page or not.

%% file: conclusion.tex
%%SL.5.20: this first sentence is very underwhelming for a conclusion section. Can we lead with something stronger and more impressive?
%We have highlighted a difficulty of reactive policy representations to generalize to new tasks, and suggested an improved policy architecture, 
%, in tasks where the approximate plan can effectively guide the reactive policy. 
%%SL.5.20: modified preceding sentence to play up the contribution

%%SL.5.20: I would cut this sentence: it puts you in a bit of a minefield with neuroscience and robotics, and doesn't really add much. Maybe just start with RL?
%Humans, over their lifetime, acquire a remarkable skill to manipulate never-seen-before objects and solve novel control problems with a single successful shot; state-of-the-art in robotics, on the other hand, is currently nowhere near these capabilities.
%%SL.5.20: be more positive: The introduction of powerful and scalable RL methods has opened up a range of new problems for deep learning. However, few recent works investigate policy architectures that are specifically tailored for planning under uncertainty. This work takes...
The introduction of powerful and scalable RL methods has opened up a range of new problems for deep learning. However, few recent works investigate policy architectures that are specifically tailored for planning under uncertainty, and current RL theory and benchmarks rarely investigate the generalization properties of a trained policy \citep{sutton1998reinforcement,mnih2015human,duan2016benchmarking}.
%While RL is a promising direction for learning robotic manipulation and control \citep{finn2016endtoend, kober2013reinforcement}, 
%%SL.5.20: one thing to note is that people *have* worked on RL architectures (see dueling networks), but from a slightly different perspective. So be careful to emphasize the planning aspect of it. I guess so far, most architectures have focused on synergy with the algorithm (e.g. dueling network for Q learning), rather than picking out something about the structure of the decision making task that is unique. In some ways, your method is to planning what convolutions are to images: maybe something worth pointing out somewhere.
This work takes a step in this direction, by exploring better generalizing policy representations. 

Our VIN policies learn an approximate planning computation relevant for solving the task, and we have shown that such a computation leads to better generalization in a diverse set of tasks, ranging from simple gridworlds that are amenable to value iteration, to continuous control, and even to navigation of Wikipedia links.
In future work we intend to learn different planning computations, based on simulation \citep{guo2014deep}, or optimal linear control \citep{watter2015embed}, and combine them with reactive policies, to potentially develop RL solutions for task and motion planning \cite{kaelbling2011hierarchical}.

%% file: appendix.tex
\normalsize
\section{Visualization of Learned Reward and Value}
\begin{figure}[ht]
\begin{center} 
    \includegraphics[width=0.315\textwidth]{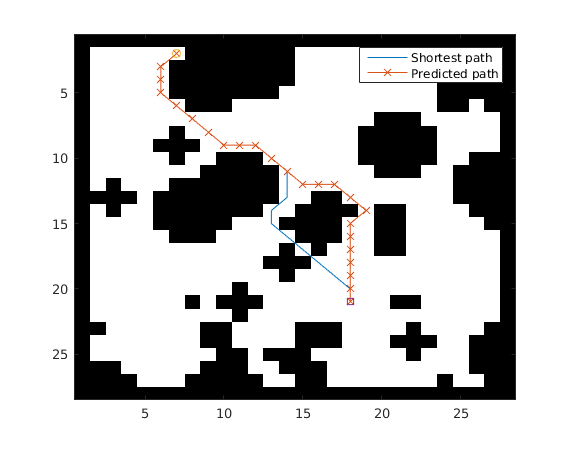}~
    \includegraphics[width=0.6\textwidth,trim={100 0 100 0},clip]{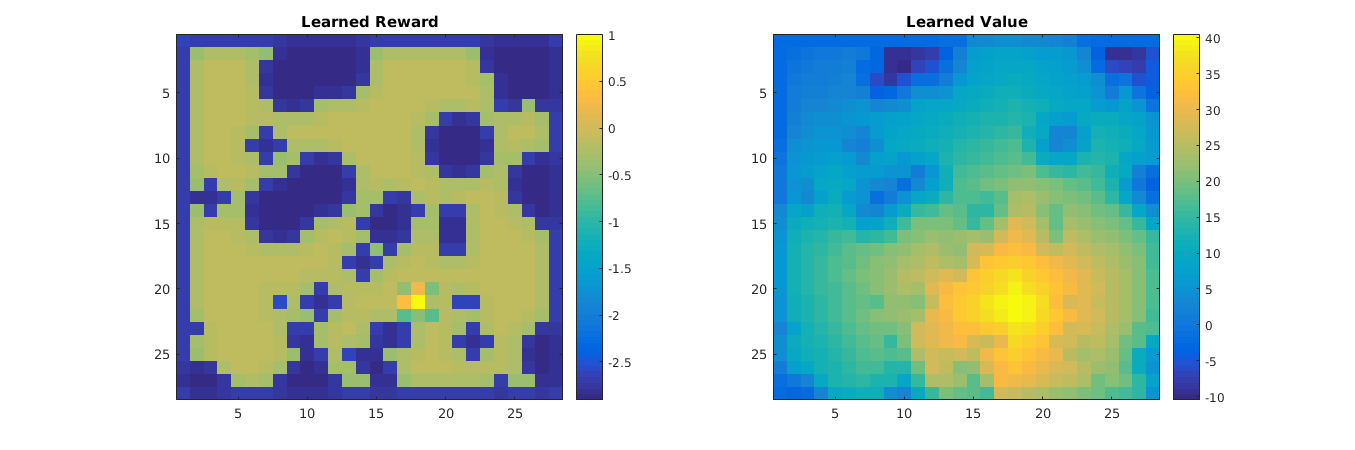}
\end{center}
\caption{Visualization of learned reward and value function. Left: a sample domain. Center: learned reward $f_R$ for this domain. Right: resulting value function (in VI block) for this domain. \label{fig:visualization}   }
\end{figure}

In Figure \ref{fig:visualization} we plot the learned reward and value function for the gridworld task. The learned reward is very negative at obstacles, very positive at goal, and a slightly negative constant otherwise. The resulting value function has a peak at the goal, and a gradient pointing towards a direction to the goal around obstacles. This plot clearly shows that the VI block learned a useful planning computation.

\section{Weight Sharing}
The VINs have an effective depth of $K$, which is larger than the depth of the reactive policies. One may wonder, whether any deep enough network would learn to plan. In principle, a CNN or FCN of depth $K$ has the potential to perform the same computation as a VIN. However, it has much more parameters, requiring much more training data. We evaluate this by untying the weights in the $K$ recurrent layers in the VIN. Our results, in Table \ref{tab:gridworld_share} show that untying the weights degrades performance, with a stronger effect for smaller sizes of training data.
\begin{table}[h]
\begin{center}
  \begin{tabular}{ | c | c | c | c | c | c | c |}
    \hline
    \multirow{2}{*}{Training data} 
    & \multicolumn{3}{|c|}{VIN} 
    & \multicolumn{3}{|c|}{VIN Untied Weights}\\ \cline{2-7}
     & Pred. & Succ. & Traj. 
     & Pred. & Succ. & Traj.\\ 
     & loss & rate & diff. 
     & loss & rate & diff.\\ \hline 

    $20\%$ & 0.06 & 98.2\% & 0.106 & 0.09 & 91.9\% & 0.094  \\ \hline 
    $50\%$ & 0.05 & 99.4\% & 0.018 & 0.07 & 95.2\% & 0.078  \\ \hline 
    $100\%$ & 0.05 & 99.3\% & 0.089 & 0.05 & 95.6\% & 0.068 \\ \hline 
  \end{tabular}
\end{center}
\caption{Performance on $16 \times 16$ grid-world domain. Evaluation of the effect of VI module shared weights relative to data size. \label{tab:gridworld_share}   }
\end{table}

\section{Gridworld with Reinforcement Learning}
We demonstrate that the value iteration network can be trained using reinforcement learning methods and achieves favorable generalization properties as compared to standard convolutional neural networks (CNNs).

The overall setup of the experiment is as follows: we train policies parameterized by VINs and policies parameterized by convolutional networks on the same set of randomly generated gridworld maps in the same way (described below) and then test their performance on a held-out set of test maps, which was generated in the same way as the set of training maps but is disjoint from the training set.

The MDP is what one would expect of a gridworld environment -- the states are the positions on the map; the actions are movements up, down, left, and right; the rewards are $+1$ for reaching the goal, $-1$ for falling into a hole, and $-0.01$ otherwise (to encourage the policy to find the shortest path); the transitions are deterministic.

\textbf{Structure of the networks.}
The VINs used are similar to those described in the main body of the paper. After $K$ value-iteration recurrences, we have approximate $Q$ values for every state and action in the map. The attention selects only those for the current state, and these are converted to a probability distribution over actions using the softmax function. We use $K = 10$ for the $8 \times 8$ maps and $K = 20$ for the $16 \times 16$ maps.

The convolutional networks' structure was adapted to accommodate the size of the maps. For the $8 \times $8 maps, we use 50 filters in the first layer and then 100 filters in the second layer, all of size $3 \times 3$. Each of these layers is followed by a $2 \times 2$ max-pool. At the end we have a fully connected hidden layer with 100 hidden units, followed by a fully-connected layer to the (4) outputs, which are converted to probabilities using the softmax function.

The network for the $16 \times 16$ maps is similar but uses three convolutional layers (with 50, 100, and 100 filters respectively), the first two of which are $2 \times 2$ max-pooled, followed by two fully-connected hidden layers (200 and 100 hidden units respectively) before connecting to the outputs and performing softmax.

\textbf{Training with a curriculum.} To ensure that the policies are not simply memorizing specific maps, we randomly select a map before each episode.  But some maps are far more difficult than others, and the agent learns best when it stands a reasonable chance of reaching this goal. Thus we found it beneficial to begin training on the easiest maps and then gradually progress to more difficult maps. This is the idea of \emph{curriculum training}.

We consider curriculum training as a way to address the exploration problem. If a completely untrained agent is dropped into a very challenging map, it moves randomly and stands approximately zero chance of reaching the goal (and thus learning a useful reward). But even a random policy can consistently reach goals nearby and learn something useful in the process, e.g. to move toward the goal. Once the policy knows how to solve tasks of difficulty $n$, it can more easily learn to solve tasks of difficulty $n+1$, as compared to a completely untrained policy. This strategy is well-aligned with how formal education is structured; you can't effectively learn calculus without knowing basic algebra.

Not all environments have an obvious difficulty metric, but fortunately the gridworld task does. We define the difficulty of a map as the length of the shortest path from the start state to the goal state. It is natural to start with difficulty 1 (the start state and goal state are adjacent) and ramp up the difficulty by one level once a certain threshold of ``success'' is reached. In our experiments we use the average discounted return to assess progress and increase the difficulty level from $n$ to $n+1$ when the average discounted return for an iteration exceeds $1 - \frac{n}{35}$. This rule was chosen empirically and takes into account the fact that higher difficulty levels are more difficult to learn.

All networks were trained using the trust region policy optimization (TRPO) \citep{schulman2015trust} algorithm, using publicly available code in the RLLab benchmark \citep{duan2016benchmarking}. 

\textbf{Testing.} When testing, we ignore the exact rewards and measure simply whether or not the agent reaches the goal. For each map in the test set, we run an episode, noting if the policy succeeds in reaching the goal. The proportion of successful trials out of all the trials is reported for each network. (See Table ~\ref{tab:gridworldRL}.)
\begin{table}
\begin{center}
  \begin{tabular}{| c | c | c |}
    \hline
    Network & $8 \times 8$ & $16 \times 16$\\
    \hline
    VIN & 90.9\% & 82.5\% \\
    \hline
    CNN & 86.9\% & 33.1\% \\
    \hline
  \end{tabular}
\end{center}
\caption{RL Results -- performance on test maps. \label{tab:gridworldRL}   }
\end{table}

\begin{figure}
\begin{center} 
    \includegraphics[width=0.4\textwidth]{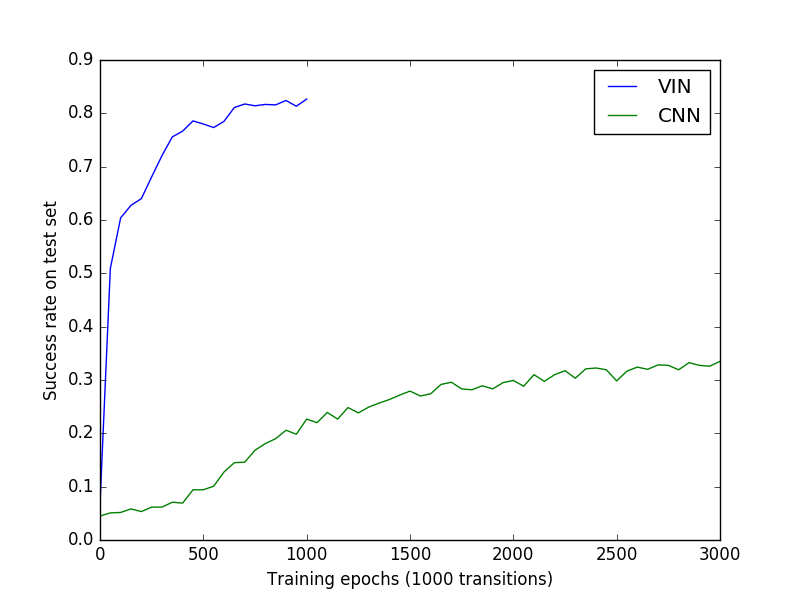}~
    \includegraphics[width=0.4\textwidth]{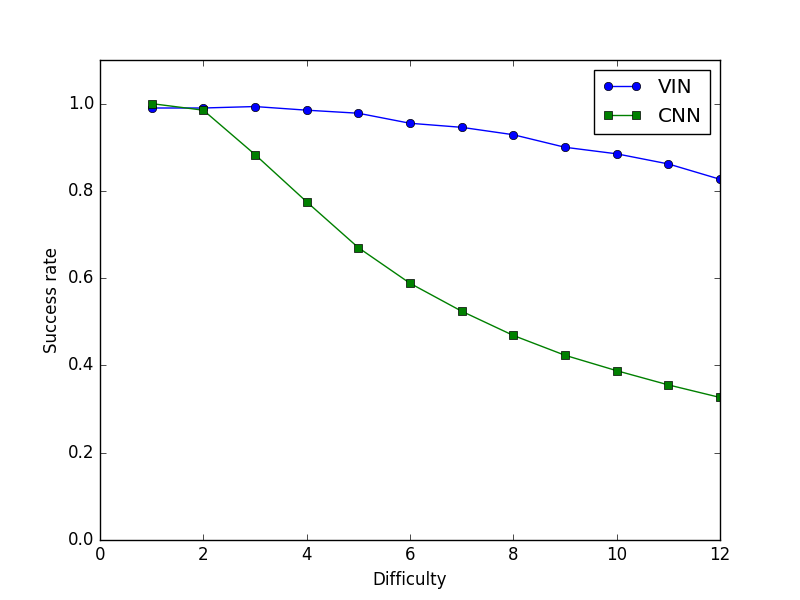}
\end{center}
\caption{RL results -- performance of VIN and CNN on $16 \times 16$ test maps. Left: Performance on all maps as a function of amount of training. Right: Success rate on test maps of increasing difficulty.}
\label{fig:rltest}
\end{figure}

On the $8 \times 8$ maps, we used the same number of training iterations on both types of networks to make the comparison as fair as possible. On the $16 \times 16$ maps, it became clear that the convolutional network was struggling, so we allowed it twice as many training iterations as the VIN, yet it still failed to achieve even a remotely similar level of performance on the test maps. (See left image of Figure \ref{fig:rltest}.) We posit that this is because the VIN learns to plan, while the CNN simply follows a reactive policy. Though the CNN policy performs reasonably well on the smaller domains, it does not scale to larger domains, while the VIN does. (See right image of Figure \ref{fig:rltest}.)

\section{Technical Details for Experiments}
We report the full technical details used for training our networks.

\subsection{Grid-world Domain}
Our training set consists of $N_i=5000$ random grid-world instances, with $N_t=7$ shortest-path trajectories (calculated using an optimal planning algorithm) from a random start-state to a random goal-state for each instance; a total of $N_i \times N_t$ trajectories. For each state $s=(i,j)$ in each trajectory, we produce a $(2 \times m \times n)$-sized observation image $s_{\text{image}}$. The first channel of $s_{\text{image}}$ encodes the obstacle presence (1 for obstacle, 0 otherwise), while the second channel encodes the goal position (1 at the goal, 0 otherwise). The full observation vector is $\phi(s) = [s, s_{\text{image}}]$. In addition, for each state we produce a label $a$ that encodes the action (one of 8 directions) that an optimal shortest-path policy would take in that state. 

We design a VIN for this task as follows. The state space $\bar{S}$ was chosen to be a $m \times n$ grid-world, similar to the true state space $S$.\footnote{For a particular configuration of obstacles, the true grid-world domain can be captured by a $m \times n$ state space with the obstacles encoded in the MDP transitions, as in our notation. For a general obstacle configuration, the obstacle positions have to also be encoded in the state. The VIN was able to learn a policy for a general obstacle configuration by planning in a $m \times n$ state space by also taking into account the observation of the map.} The reward $\bar{R}$ in this space can be represented by an $m \times n$ map, and we chose the reward mapping $f_R$ to be a CNN with $s_{\text{image}}$ as its input, one layer with $150$ kernels of size $3 \times 3$, and a second layer with one $3 \times 3$ filter to output $\bar{R}$. Thus, $f_R$ maps the image of obstacles and goal to a `reward image'. The transitions $\bar{P}$ were defined as $3 \times 3$ convolution kernels in the VI block, and exploit the fact that transitions in the grid-world are local. Note that the transitions defined this way do not depend on the state $\bar{s}$. Interestingly, we shall see that the network learned rewards and transitions that nevertheless enable it to successfully plan in this task. For the attention module, since there is a one-to-one mapping between the agent position in $S$ and in $\bar{S}$,
% since there is a one-to-one mapping between $S$ and $\bar{S}$, 
we chose a trivial approach that selects the $\bar{Q}$ values in the VI block for the state in the real MDP $s$, i.e., $\psi(s) = \bar{Q}(s,\cdot)$. The final reactive policy is a fully connected softmax output layer with weights $W$,
$\pi_{\text{re}}(\cdot|\psi(s)) \propto \exp\left( W^\top \psi(s) \right).$

We trained several neural-network policies based on a multi-class logistic regression loss function using stochastic gradient descent, with an RMSProp step size \citep{Tieleman2012}, implemented in the Theano \cite{Bastien-Theano-2012} library. 

We compare the policies:
\paragraph{VIN network}
We used the VIN model of Section \ref{s:VIN} as described above, with $10$ channels for the $q$ layer in the VI block. The recurrence $K$ was set relative to the problem size: $K=10$ for $8 \times 8$ domains, $K=20$ for $16 \times 16$ domains, and $K=36$ for $28 \times 28$ domains. The guideline for choosing these values was to keep the network small while guaranteeing that goal information can flow to every state in the map.

\textbf{CNN network:}
We devised a CNN-based reactive policy inspired by the recent impressive results of DQN \cite{mnih2015human}, with 5 convolution layers with $[50, 50, 100, 100, 100]$ kernels of size $3 \times 3$, and $2 \times 2$ max-pooling after the first and third layers. The final layer is fully connected, and maps to a softmax over actions. To represent the current state, we added to $s_{\text{image}}$ a channel that encodes the current position (1 at the current state, 0 otherwise). 

\textbf{Fully Convolutional Network (FCN):}
The problem setting for this domain is similar to semantic segmentation \citep{long2015fully}, in which each pixel in the image is assigned a semantic label (the action in our case). We therefore devised an FCN inspired by a state-of-the-art semantic segmentation algorithm \citep{long2015fully}, with 3 convolution layers, where the first layer has a filter that spans the whole image, to properly convey information from the goal to every other state.
The first convolution layer has 150 filters of size $(2m-1)\times(2n-1)$, which span the whole image and can convey information about the goal to every pixel. The second layer has 150 filters of size $1\times1$, and the third layer has 10 filters of size $1\times1$, to produce an output sized $10 \times m \times n$, similarly to the $\bar{Q}$ layer in our VIN. Similarly to the attention mechanism in the VIN, the values that correspond to the current state (pixel) are passed to a fully connected softmax output layer.

\subsection{Mars Domain}

We consider the problem of autonomously navigating the surface of Mars by a rover such as the Mars Science Laboratory (MSL) (Lockwood,
2006) over long-distance trajectories. The MSL has a limited ability for climbing high-degree slopes, and its path-planning algorithm should therefore avoid navigating into high-slope areas. In our experiment, we plan trajectories that avoid slopes of $10$ degrees or more, using overhead terrain images from the High Resolution Imaging Science Experiment (HiRISE) (McEwen et al., 2007). 
The HiRISE data consists of grayscale images of the Mars terrain, and matching elevation data,
accurate to tens of centimeters. We used an image of a $33.3$km by $6.3$km area at 49.96 degrees latitude and 219.2 degrees longitude, with a $10.5$ sq. meters / pixel resolution. 
Each domain is a $128\times 128$ image patch, on which we defined a $16 \times 16$ grid-world, where each state was considered an obstacle if its corresponding $8\times 8$ image patch contained an angle of $10$ degrees or more, evaluated using an additional elevation data. An example of the domain and terrain image is depicted in Figure \ref{fig:gridworld}. The MDP for shortest-path planning in this case is similar to the grid-world domain of Section \ref{ss:grid-world}, and the VIN design was similar, only with a deeper CNN in the reward mapping $f_R$ for processing the image. 

Our goal is to train a network that predicts the shortest-path trajectory \emph{directly from the terrain image data}. We emphasize that the ground-truth elevation data \emph{is not part of the input}, and the elevation therefore must be inferred (if needed) from the terrain image itself.

Our VIN design follows the model of Section \ref{ss:grid-world}. In this case, however, instead of feeding in the obstacle map, we feed in the raw terrain image, and accordingly modify the reward mapping $f_R$ with 2 additional CNN layers for processing the image: the first with 6 kernels of size $5 \times 5$ and $4 \times 4$ max-pooling, and the second with a 12 kernels of size $3 \times 3$ and $2 \times 2$ max-pooling. The resulting $12\times m \times n$ tensor is concatenated with the goal image, and passed to a third layer with 150 kernels of size $3 \times 3$ and a fourth layer with one $3 \times 3$ filter to output $\bar{R}$. The state inputs and output labels remain as in the grid-world experiments. We emphasize that the whole network is trained end-to-end, without pre-training the input filters.

In Table \ref{tab:mars} we present our results for training a $m=n=16$ map from a $10$K image-patch dataset, with 7 random trajectories per patch, evaluated on a held-out test set of $1$K patches. 
Figure \ref{fig:gridworld} shows an instance of the input image, the obstacles, the shortest-path trajectory, and the trajectory predicted by our method. To put the $84.8$\% success rate in context, we compare with the best performance achievable without access to the elevation data. To make this comparison, we trained a CNN to classify whether an $8 \times 8$ patch is an obstacle or not. This classifier was trained using the same image data as the VIN network, but its labels were the true obstacle classifications from the elevation map (we reiterate that the VIN network \emph{did not} have access to these ground-truth obstacle classification labels during training or testing). Training this classifier is a standard binary classification problem, and its performance represents the best obstacle identification possible with our CNN in this domain. The best-achievable shortest-path prediction is then defined as the shortest path in an obstacle map generated by this classifier from the raw image. The results of this optimal predictor are reported in Table \ref{tab:gridworld}. The 90.3\% success rate shows that obstacle identification from the raw image is indeed challenging. Thus, the success rate of the VIN network, which was trained without any obstacle labels, and had to `figure out' the planning process is quite remarkable.

\begin{table}
\begin{center}
\begin{tabular}{ | c | c | c | c | }
    \hline
    & Pred. & Succ. & Traj.\\ 
    & loss & rate & diff.\\ \hline 

    VIN & 0.089 & 84.8\% & 0.016 \\ \hline \hline
    Best  & - & 90.3\% & 0.0089 \\
    achievable & & &\\ \hline 
  \end{tabular}
  
\end{center}
\caption{Performance of VINs on the Mars domain. For comparison, the performance of a planner that used obstacle predictions trained from labeled obstacle data is shown. This upper bound on performance demonstrates the difficulty in identifying obstacles from the raw image data. Remarkably, the VIN achieved close performance \emph{without access} to any labeled data about the obstacles. \label{tab:mars}   }
\end{table}

\subsection{Continuous Control}
For training we chose the guided policy search (GPS) algorithm with unknown dynamics \citep{levine2014learning}, which is suitable for learning policies for continuous dynamics with contacts, and we used the publicly available GPS code \citep{fzftm-gpsi-16}, and Mujoco \citep{todorov2012mujoco} for physical simulation. 
GPS works by learning time-varying iLQG controllers for each domain, and then fitting the controllers to a single NN policy using supervised learning. This process is repeated for several iterations, and a special cost function is used to enforce an agreement between the trajectory distribution of the iLQG and NN controllers. We refer to \citep{levine2014learning,fzftm-gpsi-16} for the full algorithm details. For our task, we ran 10 iterations of iLQG, with the cost being a quadratic distance to the goal, followed by one iteration of NN policy fitting. This allows us to cleanly compare VINs to other policies without GPS-specific effects. 

Our VIN design is similar to the grid-world cases: the state space $\bar{S}$ is a $16 \times 16$ grid-world, and the transitions $\bar{P}$ are $3 \times 3$ convolution kernels in the VI block, similar to the grid-world of Section \ref{ss:grid-world}. However, we made some important modifications: the attention module selects a $5 \times 5$ patch of the value $\bar{V}$, centered around the current (discretized) position in the map. The final reactive policy is a 3-layer fully connected network, with a 2-dimensional continuous output for the controls. 
In addition, due to the limited number of training domains, we pre-trained the VIN with transition weights that correspond to discounted grid-world transitions
(for example, the transitions for an action to go north-west would be $\gamma$ in the top left corner and zeros otherwise), before training end-to-end. 
This is a reasonable prior for the weights in a 2-d task, and we emphasize that even with this initialization, the initial value function is meaningless, since the reward map $f_R$ is not yet learned.
The reward mapping $f_R$ is a CNN with $s_{\text{image}}$ as its input, one layer with $150$ kernels of size $3 \times 3$, and a second layer with one $3 \times 3$ filter to output $\bar{R}$. 

\subsection{WebNav}
``WebNav''~\cite{nogueira2016webnav} is a recently proposed goal-driven web navigation benchmark. In WebNav, web pages and links from some website form a directed graph $G(S,E)$. The agent is presented with a query text, which consists of $N_q$ sentences from a target page at most $N_h$ hops away from the starting page. The goal for the agent is to navigate to that target page from the starting page via clicking at most $N_n$ links per page. Here, we choose $N_h=N_q=N_n=4$. In \cite{nogueira2016webnav}, the agent receives a reward of 1 when reaching the target page via any path no longer than 10 hops. For evaluation convenience, in our experiment, the agent can receive a reward only if it reaches the destination via the \emph{shortest path}, which makes the task much harder. We measure the top-1 and top-4 prediction accuracy as well as the average reward for the baseline~\cite{nogueira2016webnav} and our VIN model.

For every page $s$, the valid transitions are $A_s=\{s' : (s, s') \in E\}$.

For every web page $s$ and every query text $q$, we utilize the bag-of-words model with pretrained word embedding provided by \cite{nogueira2016webnav} to produce feature vectors $\phi(s)$ and $\phi(q)$. The agent should choose at most $N_n$ valid actions from $A_s=\{s':(s,s')\in E\}$ based on the current $s$ and $q$.

The baseline method of \cite{nogueira2016webnav} uses a single tanh-layer neural net parametrized by $W$ to compute a hidden vector $h$:
$
h(s,q)=\textrm{tanh}\left(W\left[\begin{array}{c}
\phi(s)\\
\phi(q)
\end{array}\right]\right).
$
The final baseline policy is computed via $\pi_{\textrm{bsl}}(s'|s,q)\propto \exp \left(h(s,q)^\top \phi(s') \right)$ for $s'\in A_s$.
 
We design a VIN for this task as follows. We firstly selected a smaller website as the approximate graph $\bar{G}(\bar{S},\bar{E})$, and choose $\bar{S}$ as the states in VI. For query $q$ and a page $\bar{s}$ in $\bar{S}$, we compute the reward $\bar{R}(\bar{s})$ by $f_R(\bar{s}|q)=\textrm{tanh}\left(\left(W_R\phi(q)+b_R\right)^\top \phi(\bar{s})\right)$  with parameters $W_R$ (diagonal matrix) and $b_R$ (vector). For transition, since the graph remains unchanged, $\bar{P}$ is fixed. For the attention module $\Pi(\bar{V}^\star,s)$, we compute it by
$
\Pi(\bar{V}^\star,s)=\sum_{\bar{s}\in \bar{S}}\textrm{sigmoid}\left(\left(W_\Pi \phi(s)+b_\Pi\right)^\top\phi(\bar{s})\right)\bar{V}^\star(\bar{s}),
$
where $W_\Pi$ and $b_\Pi$ are parameters and $W_\Pi$ is diagonal. Moreover, we compute the coefficient $\gamma$ based on the query $q$ and the state $s$ using a tanh-layer neural net parametrized by $W_\gamma$:
$
\gamma(s,q)=\tanh\left(W_\gamma 
\left[
\begin{array}{c}
\phi(s)\\
\phi(q)
\end{array}
\right]\right).
$
Finally, we combine the VI module and the baseline method as our VIN model by simply adding the outputs from these two networks together.

In addition to the experiments reported in the main text, we performed experiments on the full wikipedia, using  'wikipedia for schools' as the graph for VIN planning. We report our preliminary results here.
%We experiment on two datasets: (1) a smaller dataset consisting of web pages from the 'wikipedia for schools' website\footnote{\url{http://schools-wikipedia.org/}}, as reported in the main text; and (2) the  dataset on the full wikipedia website from \cite{nogueira2016webnav}. All the methods are trained using stochastic gradient descent with an RMSProp step size.

%\textbf{Wikipedia for schools: } ``Wikipedia for Schools'' (WikiSchool) is a simplified wikipedia website designed for children in Africa, which consists of more than 6000 pages and at most 292 links per page. We used the tools from \cite{nogueira2016webnav} to generate a training set of 24273 queries (74338 training samples) and testing set of 2000 queries (6229 testing samples). 

%We select all the 172 indexing pages from WikiSchool and all the links among these pages as the approximate graph. We choose $K=3$, which is the diameter of the graph. The results are presented in Tab.~\ref{tab:wikischool}.

%\begin{table}
%\begin{center}
%  \begin{tabular}{ | c | c | c | c |}
%    \hline
%    Network & Top-1 Test Err. & Top-4 Test Err. & Avg. Reward\\
%    \hline
%    BSL &  & & \\ \hline
%    VIN & & &  \\ \hline
%  \end{tabular}
%\end{center}
%\caption{Performance on the WikiSchool dataset. \label{tab:wikischool}   }
%\end{table} 

\textbf{Full wikipedia website: } The full wikipedia dataset consists 779169 training queries (3 million training samples) and 20004 testing queries (76664 testing samples) over 4.8 million  pages with maximum 300 links per page. 

We use the whole WikiSchool website as our approximate graph and set $K=4$. In VIN, to accelerate training, we firstly only train the VI module with $K=0$. Then, we fix $\bar{R}$ obtained in the $K=0$ case and jointly train the whole model with $K=4$. The results are shown in Tab.~\ref{tab:wikifull}

\begin{table}
\begin{center}
  \begin{tabular}{ | c | c | c | c |}
    \hline
    Network & Top-1 Test Err. & Top-4 Test Err. & Avg. Reward\\
    \hline
    BSL & 52.019\% & 24.424\% & 0.27779 \\
    \hline
    VIN & 50.562\%& 26.055\%& 0.30389   \\
    \hline
  \end{tabular}
\end{center}
\caption{Performance on the full wikipedia dataset. \label{tab:wikifull}   }
\end{table} 

VIN achieves 1.5\% better prediction accuracy than the baseline. Interestingly, with only 1.5\% prediction accuracy enhancement, VIN achieves 2.5\% better success rate than the baseline: note that the agent can only success when making 4 consecutive correct predictions. This indicates the VI does provide useful high-level planning information. 

\subsection{Additional Technical Comments}
\paragraph{Runtime:} For the 2D domains, different samples from the same domain share the same VI computation, since they have the same observation.
Therefore, a single VI computation is required for samples from the same domain. Using this, and GPU code (Theano), VINs are not much slower than the baselines. For the language task, however, since Theano doesn’t support convolutions on graphs nor sparse operations on GPU, VINs were considerably slower in our implementation. 
% We are currently working on a CUDA implementation, as part of our efforts to extend our results to the full Wikipedia website.

\section{Hierarchical VI Modules}
The number of VI iterations $K$ required in the VIN depends on the problem size. Consider, for example, a grid-world in which the goal is located $L$ steps away from some state $s$. Then, at least $L$ iterations of VI are required to convey the reward information from the goal to state $s$, and clearly, any action prediction obtained with less than $L$ VI iterations at state $s$ is unaware of the goal location, and therefore unacceptable.

\begin{figure}[h]
\begin{center}
\includegraphics[scale=0.48,clip=true,trim=180 180 110 50]{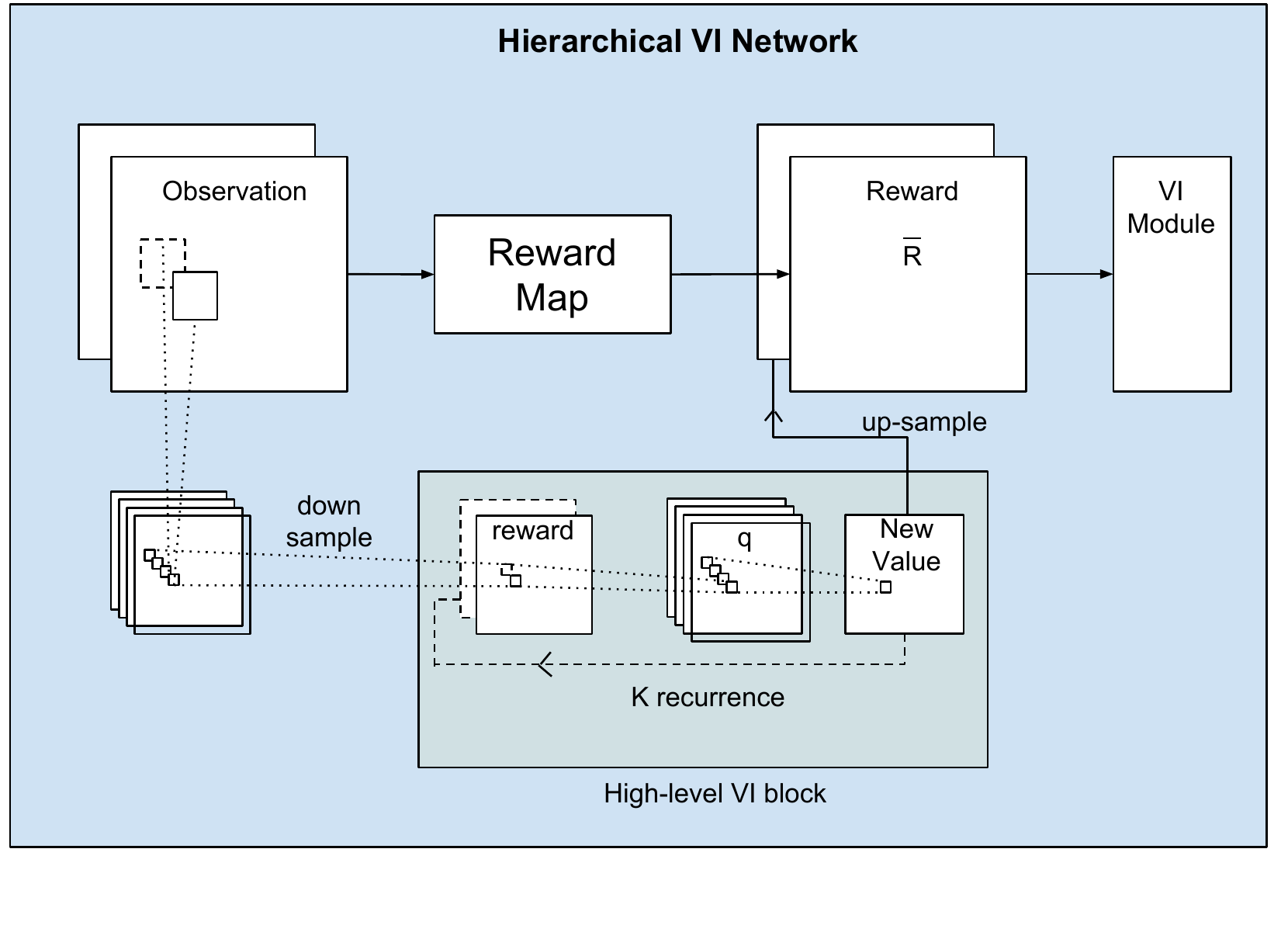}
\end{center}
\caption{Hierarchical VI network. A copy of the input is first fed into a convolution layer and then downsampled. This signal is then fed into a VI module to produce a coarse value function, corresponding to the upper level in the hierarchy. This value function is then up-sampled, and added as an additional channel in the reward layer of a standard VI module (lower level of the hierarchy). \label{fig:hvin}   }
\end{figure}

To convey reward information faster in VI, and reduce the effective $K$, we propose to perform VI at multiple levels of resolution. We term this model a hierarchical VI Network (HVIN), due to its similarity with hierarchical planning algorithms. In a HVIN, a copy of the input down-sampled by a factor of $d$ is first fed into a VI module termed the high-level VI module. The down-sampling offers a $d \times $ speedup of information transmission in the map, at the price of reduced accuracy. The value layer of the high-level VI module is then up-sampled, and added as an additional input channel to the input of the standard VI module. Thus, the high-level VI module learns a mapping from down-sampled image features to a suitable reward-shaping for the nominal VI module. The full HVIN model is depicted in Figure \ref{fig:hvin}. This model can easily be extended to include multiple levels of hierarchy.

Table \ref{tab:hvin} shows the performance of the HVIN module in the grid-world task, compared to the VIN results reported in the main text. 
We used a $2 \times 2$ down-sampling layer. Similarly to the standard VIN, $3 \times 3$ convolution kernels, $150$ channels for each hidden layer $H$ (for both the down-sampled image, and standard image), and $10$ channels for the $q$ layer in each VI block. Similarly to the VIN networks, the recurrence $K$ was set relative to the problem size, taking into account the down-sampling factor: $K=4$ for $8 \times 8$ domains, $K=10$ for $16 \times 16$ domains, and $K=16$ for $28 \times 28$ domains (in comparison, the respective $K$ values for standard VINs were $10$, $20$, and $36$). The HVINs demonstrated better performance for the larger $28\times 28$ map, which we attribute to the improved information transmission in the hierarchical VI module.

\begin{table*}
\begin{center}
  \begin{tabular}{ | c | c | c | c | c | c | c | c | c | c |}
    \hline
    \multirow{2}{*}{Domain} & \multicolumn{3}{|c|}{VIN} & \multicolumn{3}{|c|}{Hierarchical VIN} \\ \cline{2-7}
     & Prediction & Success & Trajectory & Prediction & Success & Trajectory \\ 
     & loss & rate & diff. & loss & rate & diff. \\ \hline 

    $8 \times 8$ & 0.004 & \bf 99.6\% & 0.001 & 0.005 & 99.3\% & 0.0 \\ \hline 
    $16 \times 16$ & 0.05 & \bf 99.3\% & 0.089 &  0.03 & 99\% & 0.007 \\ \hline 
    $28 \times 28$ & 0.11 & 97\% & 0.086 & 0.05 & \bf 98.1\% & 0.037 \\ \hline 
  \end{tabular}
 
\end{center}
\caption{HVIN performance on grid-world domain.  \label{tab:hvin}   }
\end{table*}